\documentclass[journal]{IEEEtran}
\usepackage[table]{xcolor}
\usepackage{graphicx,amsmath,amssymb,epstopdf,mathtools,multirow,booktabs}
\usepackage{subcaption}	
\usepackage{amsthm,float}
\usepackage{threeparttable, tablefootnote}
\usepackage{algpseudocode}
\usepackage{arydshln}
\usepackage{pdfrender}

\usepackage[hidelinks]{hyperref}
\usepackage{paralist}

\floatstyle{ruled}
\newfloat{algorithm}{tbp}{loa}
\providecommand{\algorithmname}{Algorithm}
\floatname{algorithm}{\protect\algorithmname}

\let\vec\boldvec

\DeclareMathOperator*{\argmin}{argmin}

\newcommand{\trsp}{{\!\scriptscriptstyle\top}}

\newcommand*{\boldcheckmark}{%
	\textpdfrender{
		TextRenderingMode=FillStroke,
		LineWidth=.5pt, 
	}{\checkmark}%
}

\definecolor{Light1}{rgb}{0.98, 0.95, 0.90}
\definecolor{Light2}{rgb}{0.98, 0.98, 0.93}
\definecolor{Light3}{rgb}{0.98, 0.98, 1}

\begin{document}
\title{Towards Orientation Learning and Adaptation in Cartesian Space}

\author{Yanlong Huang, Fares J. Abu-Dakka, Jo\~{a}o Silv\'{e}rio, and Darwin G. Caldwell 

\thanks{Yanlong Huang is with School of Computing, University of Leeds, Leeds LS29JT, UK (e-mail:  y.l.huang@leeds.ac.uk).}		
\thanks{Fares J. Abu-Dakka is with Intelligent Robotics Group at the Department of Electrical Engineering and Automation, Aalto University, Finland (e-mail:  fares.abu-dakka@aalto.fi).}
\thanks{Jo\~{a}o Silv\'{e}rio is with Idiap Research Institute, CH-1920 Martigny,
		Switzerland (e-mail: joao.silverio@idiap.ch).}
\thanks{Darwin G. Caldwell is with Department of Advanced Robotics, Istituto Italiano di Tecnologia,	Via Morego 30, 16163 Genoa, Italy (e-mail:  darwin.caldwell@iit.it).}		

\thanks{Preliminary results have been presented in IEEE International Conference on Robotics and Automation  \cite{Huang2019_1}.}}

\markboth{IEEE Transactions on Robotics}%
{IEEE Transactions on Robotics}
\maketitle

\begin{abstract}
As a promising branch of robotics, imitation learning emerges as an important way to transfer human skills to robots, where human demonstrations represented in Cartesian or joint spaces are utilized to estimate task/skill models that can be subsequently generalized to new situations. 
While learning Cartesian positions suffices for many applications, the end-effector orientation is required in many others. 
Despite recent advances in learning orientations from demonstrations, 
several crucial issues 
have not been adequately addressed yet. For instance, how can demonstrated orientations be adapted to pass through arbitrary desired points that comprise orientations and angular velocities? 
In this paper, we propose an approach that is capable of learning multiple orientation trajectories and adapting learned orientation skills to new situations (e.g., via-points and end-points), where both orientation and angular velocity are considered. Specifically, we introduce a kernelized treatment to alleviate explicit basis functions when learning orientations, which allows for learning orientation trajectories associated with high-dimensional inputs. In addition, we extend our approach to the learning of quaternions with angular acceleration or jerk constraints, which allows for generating smoother orientation profiles for robots. Several examples including experiments with real 7-DoF robot arms are provided to verify the effectiveness of our method. 

\end{abstract}

\begin{IEEEkeywords}
imitation learning,	orientation learning, generalization, human-robot collaboration.
\end{IEEEkeywords}

\section{Introduction}
\IEEEPARstart{I}n many challenging tasks (e.g., robot table tennis \cite{Huang2015} and bimanual manipulation \cite{Joao}), it is non-trivial to manually define proper trajectories for robots beforehand, hence imitation learning is suggested in order to facilitate the transfer of human skills to robots \cite{Schaal}.
The basic idea of imitation learning is to model consistent or important motion patterns that underlie human skills and, subsequently, employ these patterns in new situations. A myriad of imitation learning techniques have been reported in the past few years, such as dynamic movement primitives (DMP) \cite{Ijspeert}, probabilistic movement primitives (ProMP) \cite{Paraschos}, task-parameterized Gaussian mixture model (TP-GMM) \cite{Calinon2016} and kernelized movement primitives (KMP) \cite{Huang2017}. 

While the aforementioned skill learning approaches have been proven effective for robot trajectory generation at the level of Cartesian positions and joint angles \cite{Koert, Zhou, Joao2019}, learning of orientation in task space still imposes great challenges. Unlike position operations in Euclidean space, orientation is accompanied by additional constraints, e.g., the unit norm of the quaternion representation or the orthogonal constraint of rotation matrices. 
In many previous work, quaternion trajectories are learned and adapted
without considering the unit norm constraint (e.g., orientation TP-GMM \cite{Joao} and DMP \cite{Pastor}), leading to improper quaternions and hence requiring an additional renormalization.

\begin{table*}[bt]
	\caption {Comparison Among the State-of-the-Art	and Our Approach}
	\centering
	\scalebox{0.79}{\begin{tabular}{lcccccccc}
			\toprule %
			&Probabilistic & Unit norm & Via-quaternion & Via-angular velocity & End-quaternion & End-angular velocity & Angular acc (or jerk) constraints & Multiple inputs$^{\dagger}$
			\\ 
			\toprule %
			Silv\'{e}rio \emph{et al.} \cite{Joao}
			& \boldcheckmark
			& \textbf{-}
			& \textbf{-}
			& \textbf{-}
			& \boldcheckmark
			& \textbf{-}
			& \textbf{-}
			& \boldcheckmark
			\\
			\midrule
			Pastor \emph{et al.} \cite{Pastor11} 
			& \textbf{-}
			& \boldcheckmark 
			& \textbf{-} 
			& \textbf{-}
			& \boldcheckmark
			& \textbf{-}$^{*}$
			& \textbf{-}
			& \textbf{-}
			\\
			\midrule			
			Ude \emph{et al.} \cite{Ude}
			& \textbf{-}
			& \boldcheckmark 
			& \textbf{-}
			& \textbf{-}
			& \boldcheckmark
			& -$^{*}$
			& \textbf{-}
			& \textbf{-}
			\\
			\midrule			
			Abu-Dakka \emph{et al.} \cite{Abu}
			& \textbf{-}
			& \boldcheckmark 
			& \textbf{-}
			& \textbf{-}
			& \boldcheckmark
			& \textbf{-}$^{*}$
			& \textbf{-}
			& \textbf{-}
			\\
			\midrule
			Kim \emph{et al.} \cite{Kim}
			& \boldcheckmark 
			& \boldcheckmark
			& \textbf{-}
			& \textbf{-}
			& \textbf{-}
			& \textbf{-}
			& \textbf{-}
			& \boldcheckmark
			\\
			\midrule
			Zeestraten \emph{et al.} \cite{Matin}
			& \boldcheckmark 
			& \boldcheckmark
			& \textbf{-}
			& \textbf{-}
			& \boldcheckmark
			& \textbf{-}
			& \textbf{-}
			& \boldcheckmark
			\\
			\midrule			
			Saveriano \emph{et al.} \cite{Saveriano}
			& \textbf{-}
			& \boldcheckmark 
			& \boldcheckmark
			& \boldcheckmark
			& \boldcheckmark
			& \boldcheckmark
			& \textbf{-}
			& \textbf{-}
			\\
			\midrule			
			Kramberger \emph{et al.} \cite{Kramberger}
			& \textbf{-}
			& \boldcheckmark 
			& \textbf{-}
			& \textbf{-}
			& \boldcheckmark
			& \textbf{-}$^{*}$
			& \textbf{-}
			& \textbf{-}$^{\ddagger}$
			\\
			\midrule
			Our approach
			& \boldcheckmark
			& \boldcheckmark 
			& \boldcheckmark
			& \boldcheckmark
			& \boldcheckmark
			& \boldcheckmark
			& \boldcheckmark
			& \boldcheckmark
			\\
			\bottomrule
	\end{tabular}}
	\begin{tablenotes}
		\item[] {* \hspace{0.0cm}In these works, primitives end with zero angular velocity, i.e., one can not set a desired non-zero velocity.}
		\item[] {$\dagger$\hspace{0.0cm} The multiple inputs are parts of the demonstrations, which shall be distinguished from external contextual variables. For example, in a human-robot handover task, a typical demonstration consists of a varying human hand trajectory (i.e., high-dimensional inputs) and a robot trajectory (i.e., outputs).}
		\item[] {\hspace{-0.00cm}$\ddagger$ This work considers the generalization of demonstrations towards different external contextual (or conditional) states (which could be high-dimensional). However, the demonstrations are composed of time sequences (i.e., 1-D input) and the corresponding robot trajectories (i.e., outputs).}
	\end{tablenotes}
	\label{table:comp:table}
\end{table*}

Instead of learning quaternions in Euclidean space, a few approaches that comply with orientation constraints have been proposed. One representative type of approach is built on DMP \cite{Pastor11},$ \, $ \cite{Ude, Abu}, where unit quaternions were used to represent orientation and different reformulations of DMP were developed to ensure proper quaternions over the course of orientation adaptation. However, \cite{Pastor11},$ \, $ \cite{Ude, Abu} can only adapt quaternions towards a desired target with zero angular velocity as a consequence of the spring-damper dynamics inherited from the original DMP. 

Another solution of learning orientation was proposed in \cite{Kim}, where GMM was employed to model the distribution of quaternion displacements so as to avoid the quaternion constraint. However, this approach only focuses on orientation reproduction without addressing the adaptation issue. In contrast to \cite{Kim} that learns quaternion displacements, 
the Riemannian topology of the $\mathbb{S}^3$ manifold was exploited in \cite{Matin} to probabilistically encode and reproduce distributions of quaternions. Moreover, an extension to task-parameterized movements was provided in \cite{Matin}, which 
allows for adapting orientation tasks to different initial and final orientations. However, adaptation to orientation via-points and angular velocities is not provided.

In addition to the above-mentioned issues, learning orientations associated with high-dimensional inputs is important. 
For example, in a human-robot collaboration scenario, the robot end-effector orientation is often required to react promptly and properly according to the user's state (e.g., hand poses). More specifically, the robot might need to adapt its orientation in accordance to dynamic environments. The results \cite{Pastor,Pastor11,Ude,Abu} are built on time-driven\footnote{Despite that time is often transformed into a phase variable in DMP, we will refer to DMP as a time-driven approach, since time and phase are 1-dimensional and their mapping is bijective, which make them be equivalent in our argument.} DMP, and hence it is non-straightforward to extend these works to learn demonstrations consisting of high-dimensional varying inputs and deal with tasks where the robot should react to a high number of input variables, e.g., the hand position of a human partner in a collaborative task. In contrast, due to the employment of GMM, learning orientations with multiple inputs is feasible in \cite{Joao, Kim, Matin}. However, extending these approaches to tackle adaptations towards via-points associated with multi-dimensional inputs is non-trivial.

While many imitation learning approaches focus on mimicking human demonstrations, the constrained skill learning is often overlooked. As discussed in \cite{Koc,Ratliff}, trajectory smoothness (e.g., acceleration and jerk) will influence robot performance, particularly in time-contact systems (e.g., striking movement in robot table tennis). Thus, it is desirable to incorporate smoothness constraints into the process of learning orientations.

In summary, 
if we consider the problem of adapting quaternions and angular velocities to pass through arbitrary desired points (e.g., via-point and end-point) while taking into account  high-dimensional inputs and smoothness constraints, no previous work in the scope of imitation learning provides an all-encompassing solution.

In this paper, we aim at providing an \emph{analytical} solution that is capable of
\begin{enumerate}
\item[(\emph{i})] learning \emph{multiple} quaternion trajectories,
\item[(\emph{ii})] allowing for orientation \emph{adaptations} towards arbitrary desired points that consist of both \emph{unit quaternions and angular velocities},
\item[(\emph{iii})] coping with orientation learning and adaptations associated with  \emph{high-dimensional inputs},
\item[(\emph{iv})] accounting for  \emph{smoothness constraints}.
\end{enumerate}
For the purpose of clear comparison, the main contributions of the state-of-the-art approaches and our approach are summarized in Table~\ref{table:comp:table}. 

This paper is structured as follows. We first illustrate the probabilistic learning of multiple quaternion trajectories and derive our main results in Section~\ref{sec:model:demo}. Subsequently, we extend the obtained results to quaternion adaptations in Section~\ref{sec:ada}, as well as quaternion learning and adaptation with angular acceleration (or jerk) constraints in Section~\ref{sec:quat:const}. After that,  we take a typical human-robot collaboration case as an example to show how our approach can be applied to the learning of quaternions along with multiple inputs in Section~\ref{sec:ori:highDim}. We evaluate our method through several simulated examples (including discrete and rhythmic quaternion profiles) and real experiments (a painting task with time input on Barrett WAM robot and a handover task with a multi-dimensional input  on KUKA robot) in Section~\ref{sec:eva}. In Section~\ref{sec:discuss}, we discuss the related work as well as limitations and possible extensions of our approach. Finally, our work is concluded in Section~\ref{sec:conclusion}. 
Note that this paper has comprehensively extended our previous work \cite{Huang2019_1} in terms of both theoretical parts (e.g., Sections~\ref{sec:quat:const} and \ref{sec:ori:highDim}) and evaluations (e.g., Sections~\ref{subsec:jerk:ada:simu}, \ref{subsec:eva:rhythmic:ada:simu} and \ref{sec:real:eva:highD}).

\begin{table*}[htbp]\caption{Description of Key Notations}
	\centering 
	\begin{tabular}{l l p{0.65\textwidth} }
		\toprule
		\rowcolor{Light1}
		$\vec{q}$, $\bar{\vec{q}}$ & $\triangleq$ & quaternion and its conjugation\\
		\rowcolor{Light1}
		$\vec{q}_{a}$ & $\triangleq$ & auxiliary quaternion\\
		\rowcolor{Light1}
		${\vec{\zeta}}$& $\triangleq$ & transformed state of quaternion\\
		\rowcolor{Light1}
		${\vec{\omega}}$ & $\triangleq$ & angular velocity\\
		\rowcolor{Light1}
		$\vec{p}$ & $\triangleq$ & Cartesian position\\	
		\rowcolor{Light1}
		$C$ & $\triangleq$ & number of Gaussian components in GMM\\				
		\rowcolor{Light1}
		$\pi_c, \vec{\mu}_c, \vec{\Sigma}_c$ & $\triangleq$ & parameters of $c$--th Gaussian component in GMM, see (\ref{equ:gmm})\\
		\rowcolor{Light1}
		$\vec{w}$ & $\triangleq$ & unknown parametric vector \\
		\rowcolor{Light1}
		$\vec{\phi}(t)$, $\vec{\Theta}(t)$  & $\triangleq$ & $B$-dimensional basis function vector and its corresponding expanded matrix, see (\ref{equ:para:traj})\\
		\rowcolor{Light1}
		$\vec{\varphi}(t)$, $\vec{\Omega}(t)$ 
		& $\triangleq$ & expanded matrices, see (\ref{equ:const:notation}) and (\ref{equ:notation:add:acc})\\
		\rowcolor{Light1}
		${k}(\cdot,\cdot)$ & $\triangleq$ & kernel function\\						
		\rowcolor{Light2}							
		$\vec{D}_q=\{\{t_{n,m}, \vec{q}_{n,m}\}_{n=1}^{N}\}_{m=1}^{M}$ & $\triangleq$ & $M$ demonstrations in terms of time and quaternion, where each demonstration has $N$ datapoints\\
		\rowcolor{Light2}
		${\vec{D}_\zeta=\{\{t_{n,m}, \vec{\zeta}_{n,m}, \dot{\vec{\zeta}}_{n,m}\}_{n=1}^{N}\}_{m=1}^{M}}$ & $\triangleq$ & transformed data obtained from $\vec{D}_q$, where $\vec{\zeta}_{n,m}$ = $\log(\vec{q}_{n,m}*\bar{\vec{q}}_{a})$, see (\ref{equ:transfer})\\
		\rowcolor{Light2}
		$\vec{D}_\eta=\{\{t_{n,m}, \vec{\eta}_{n,m}\}_{n=1}^{N}\}_{m=1}^{M}$  & $\triangleq$ &  compact form of $\vec{D}_\zeta$, where 		$\vec{\eta}_{n,m} = [\vec{\zeta}_{n,m}^{\trsp}\,\dot{\vec{\zeta}}_{n,m}^{\trsp}]^{\trsp}$ \\
		\rowcolor{Light2}
		$\vec{D}_r=\{t_n,\hat{\vec{\mu}}_n,\hat{\vec{\Sigma}}_n\}_{n=1}^{N}$ & $\triangleq$ & probabilistic reference trajectory extracted from  $\vec{D}_{\eta}$, with $\hat{\vec{\eta}}_n|t_n \sim \mathcal{N}(\hat{\vec{\mu}}_n,\hat{\vec{\Sigma}}_n)$\\
		
		\rowcolor{Light2}
		$\vec{\Phi},\vec{\Sigma},\vec{\mu}$ & $\triangleq$ & expanded matrices/vectors defined on $\vec{D}_r$, see (\ref{equ:mat})\\
		\rowcolor{Light2}
		$\vec{k}(t_i,t_j)$  & $\triangleq$ & expanded kernel matrix, see (\ref{equ:kernel}) or (\ref{equ:kernel:const})\\
		\rowcolor{Light2}
		$\tilde{\vec{D}}_q=\{\tilde{t}_h,\tilde{\vec{q}}_h,\tilde{\vec{\omega}}_h\}_{h=1}^{H}$ & $\triangleq$ & $H$ desired quaternion states\\
		\rowcolor{Light2}
		$\tilde{\vec{D}}_{\zeta}=\{\tilde{t}_h,\tilde{\vec{\zeta}}_h,\dot{\tilde{\vec{\zeta}}}_h\}_{h=1}^{H}$  & $\triangleq$ & transformed states obtained from $\tilde{\vec{D}}_q$, see (\ref{equ:des:quat:transfer}) and (\ref{equ:des:der:tranfer})\\
		\rowcolor{Light2}
		$\tilde{\vec{D}}_{\eta}=\{\tilde{t}_h,\tilde{\vec{\eta}}_h\}_{h=1}^{H}$ & $\triangleq$ & compact form of $\tilde{\vec{D}}_{\zeta}$, where $\tilde{\vec{\eta}}=[\tilde{\vec{\zeta}}^{\trsp}\,\dot{\tilde{\vec{\zeta}}}^{\trsp}]^{\trsp}$\\
		\rowcolor{Light2}
		$\tilde{\vec{D}}_{r}=\{\tilde{t}_h,\tilde{\vec{\eta}}_h,\tilde{\vec{\Sigma}}_h\}_{h=1}^{H}$ & $\triangleq$ & additional reference trajectory to indicate the transformed desired points \\
		\rowcolor{Light2}
		${\vec{D}}_{r}^{U}=\{{t}_l^{U},{\vec{\mu}}_l^{U},{\vec{\Sigma}}_l^{U}\}_{l=1}^{N+H}$ & $\triangleq$ & extended reference trajectory, see (\ref{equ:ref:update})\\
		
		\rowcolor{Light3}
		$\vec{D}^0=\{\{\vec{s}_{n,m}, \vec{\xi}_{n,m}^0\}_{n=1}^{N}\}_{m=1}^{M}$ & $\triangleq$ & demonstration database with high-dimensional input $\vec{s}_{n,m}$
		and output $\vec{\xi}_{n,m}^{0}=\left[\begin{matrix}
		\vec{p}_{n,m} \\ \vec{q}_{n,m}
		\end{matrix}\right] $\\
		\rowcolor{Light3}
		$\vec{D}^s=\{\{\vec{s}_{n,m}, \vec{\xi}_{n,m}\}_{n=1}^{N}\}_{m=1}^{M}$ & $\triangleq$ & transformed data obtained from $\vec{D}^0$ with 
		$\vec{\xi}_{n,m}=\left[\begin{matrix}
		\vec{p}_{n,m} \\ \log(\vec{q}_{n,m}*\bar{\vec{q}}_{a})
		\end{matrix}\right]$ \\
		\rowcolor{Light3}
		$\vec{D}_r^s=\{\vec{s}_n,\hat{\vec{\mu}}_n,\hat{\vec{\Sigma}}_n\}_{n=1}^{N}$ & $\triangleq$ & probabilistic reference trajectory extracted from $\vec{D}^s$\\
		\rowcolor{Light3}
		$\vec{\phi}(\vec{s})$, $\vec{\Theta}(\vec{s})$  & $\triangleq$ & basis function vector with high-dimensional inputs and its corresponding expanded matrix, see (\ref{equ:para:traj:high})\\
		\rowcolor{Light3}
		$\vec{k}(\vec{s}_i,\vec{s}_j)$  & $\triangleq$ & expanded kernel matrix, see (\ref{equ:kernel:high})\\
		\rowcolor{Light3}
		$\tilde{\vec{D}}^{0}=\{\tilde{\vec{s}}_h,\tilde{\vec{\xi}}_h^{0}\}_{h=1}^{H}$ & $\triangleq$ & $H$ desired points associated with high-dimensional inputs\\
		\rowcolor{Light3}
		$\tilde{\vec{D}}^s=\{\tilde{\vec{s}}_h,\tilde{\vec{\xi}}_h\}_{h=1}^{H}$ & $\triangleq$ & transformed desired data from $\tilde{\vec{D}}^{0}$\\
		\rowcolor{Light3}
		$\tilde{\vec{D}}_r^s=\{\tilde{\vec{s}}_h,\tilde{\vec{\xi}}_h,\tilde{\vec{\Sigma}}_h\}_{h=1}^{H}$ & $\triangleq$ & additional reference trajectory for high-dimensional inputs\\
		
		\bottomrule
	\end{tabular}
	\label{tab:notation}
\end{table*}

\section{Probabilistic Learning of Quaternion Trajectories \label{sec:model:demo}}

As suggested in \cite{Calinon2016, Muhlig}, the probability distribution of multiple demonstrations often encapsulates important motion features and further facilitates the design of optimal controllers \cite{Joao2019, Todorov, Medina, Huang2018}. Nonetheless, 
the direct probabilistic modeling of quaternion trajectories is intractable as a result of the unit norm constraint. Similarly to \cite{Ude, Kim, Matin}, we propose to transform quaternions into Euclidean space, which hence enables the probabilistic modeling of transformed trajectories (Section~\ref{subsec:gmm}). Then, we exploit the distribution of transformed trajectories using a kernelized approach,
whose predictions allow for the retrieval of proper quaternions (Section~\ref{subsec:kmp}). 
We summarize key notations used throughout the paper in Table~\ref{tab:notation}.

\subsection{Probabilistic modeling of quaternion trajectories\label{subsec:gmm}}

For the sake of clarity, 
let us define quaternions $\vec{q}_1=
\begin{bmatrix} v_1 \\ \vec{u}_1 \end{bmatrix}
$ and $\vec{q}_2=
\begin{bmatrix} v_2 \\ \vec{u}_2 \end{bmatrix}
$, where $\vec{q}_i \in \mathbb{S}^3$, $v_i\in\mathbb{R}$ and $\vec{u}_i\in\mathbb{R}^3$, $i\in\{1,2\}$. Besides, we write $\bar{\vec{q}}_2=
\begin{bmatrix} v_2 \\ -\vec{u}_2 \end{bmatrix}
$ as the conjugation of $\vec{q}_2$ and, $\vec{q}=\vec{q}_1*\bar{\vec{q}}_2=
\begin{bmatrix} v \\ \vec{u} \end{bmatrix}
$ as the quaternion product
of $\vec{q}_1$ and $\bar{\vec{q}}_2$.
The function $\log(\cdot): \mathbb{S}^3 \mapsto \mathbb{R}^3$ that can be used to determine the difference vector between $\vec{q}_1$ and $\vec{q}_2$ is defined as \cite{Ude}
\begin{equation}
\log(\vec{q}_1*\bar{\vec{q}}_2)=\log(\vec{q})=
\left\{ 
\begin{aligned}
&\arccos(v) \frac{\vec{u}}{||\vec{u}||}, \vec{u}\neq \vec{0} \\
&[0\;\; 0\;\; 0]^{\trsp}, \quad\quad otherwise,
\end{aligned}  
\right.
\label{equ:quat:diff}
\end{equation}
where $||\cdot||$ denotes $\ell_2$ norm.
By using this function, demonstrated quaternions can be projected into Euclidean space.

Let us assume that we can access a set of demonstrations $\vec{D}_q=\{\{t_{n,m}, \vec{q}_{n,m}\}_{n=1}^{N}\}_{m=1}^{M}$ with $N$ being the time length and $M$ the number of demonstrations, where $\vec{q}_{n,m}$ denotes a quaternion at the $n$-th time-step from the $m$-th demonstration. 
In addition, we introduce an auxiliary quaternion\footnote{$\vec{q}_{a}$ should meet the constraint: $\vec{q}_{n,m}*\bar{\vec{q}}_{a} \neq [-1 \ 0 \ 0 \ 0]^{\trsp}$, which shall be seen in the Assumption 2 explained later.} $\vec{q}_{a}$, 
which is subsequently used for transforming demonstrated quaternions into Euclidean space, yielding new trajectories as ${\vec{D}_\zeta=\{\{t_{n,m}, \vec{\zeta}_{n,m}, \dot{\vec{\zeta}}_{n,m}\}_{n=1}^{N}\}_{m=1}^{M}}$ with 
\begin{equation}
{\vec{\zeta}_{n,m}=\log(\vec{q}_{n,m}*\bar{\vec{q}}_{a}) }
\label{equ:transfer}
\end{equation}
and $\dot{\vec{\zeta}}_{n,m}\in \mathbb{R}^3$ being the derivative of $\vec{\zeta}_{n,m}\in \mathbb{R}^3$.

It is worth pointing out that $\vec{q}$ and $-\vec{q}$ denote the same orientation. In order to ensure that all demonstrations have no discontinuities, we assume that $\vec{q}_{n,m}^{\trsp}\vec{q}_{n+1,m}>0$, $\forall n \in \{1,2,\ldots,N-1\}$, $\forall m\in\{1,2,\ldots,M\}$. Note that if this is not satisfied, we can simply multiply $\vec{q}_{n+1,m}$ by $-1$.
In addition, from the definition of $\log(\cdot)$ in (\ref{equ:quat:diff}), we can see that $\log(\vec{q}*\bar{\vec{q}}_{a})$ and $\log(-\vec{q}*\bar{\vec{q}}_{a})$ are different, albeit that $\vec{q}*\bar{\vec{q}}_{a}$ and $-\vec{q}*\bar{\vec{q}}_{a}$ represent the same orientation. To avoid this issue, at the $n$-th time step with $n\in\{1,2,\ldots,N\}$, we assume 
$(\vec{q}_{n,i}*\bar{\vec{q}}_{a})^{\trsp}(\vec{q}_{n,j}*\bar{\vec{q}}_{a})>0$, $\forall i,j \in \{1,2,\ldots,M\}$, implying that
$\vec{q}_{n,1}*\bar{\vec{q}}_{a},\vec{q}_{n,2}*\bar{\vec{q}}_{a},\ldots,\vec{q}_{n,M}*\bar{\vec{q}}_{a}$ stay in the same hemisphere of $\mathbb{S}^3$. 
If we write quaternion product into the form of matrix-vector multiplication, we have
$(\vec{q}_{n,i}*\bar{\vec{q}}_{a})^{\trsp}(\vec{q}_{n,j}*\bar{\vec{q}}_{a})=\bigl(\vec{A}(\bar{\vec{q}}_{a}) \vec{q}_{n,i}\bigr)^{\trsp}\bigl(\vec{A}(\bar{\vec{q}}_{a}) \vec{q}_{n,j}\bigr)=\vec{q}_{n,i}^{\trsp}\vec{q}_{n,j}$,
where $\vec{A}(\bar{\vec{q}}_{a}) \in \mathbb{R}^{4\times 4}$ is an orthogonal matrix \cite{Joao}.
In summary, demonstrations $\vec{D}_q$ should satisfy the following assumption:

\textbf{{Assumption 1}} \emph{$\vec{q}_{n,i}^{\trsp}\vec{q}_{n,j}>0$, $\forall n\in \{1,2,\ldots,N\}$, $\forall i,j \in \{1,2,\ldots,M\}$.
Moreover, $\vec{q}_{n,m}^{\trsp}\vec{q}_{n+1,m}>0$,  $\forall n\in \{1,2,\ldots,N-1\}$, $\forall m \in \{1,2,\ldots,M\}$.}

For simplicity, we denote  $\vec{\eta}=[\vec{\zeta}^{\trsp}\,\dot{\vec{\zeta}}^{\trsp}]^{\trsp} \in \mathbb{R}^6$ and accordingly $\vec{D}_\zeta$ becomes ${\vec{D}_\eta=\{\{t_{n,m}, \vec{\eta}_{n,m}\}_{n=1}^{N}\}_{m=1}^{M}}$. Now, we apply GMM  \cite{Calinon2016} to model the joint probability distribution $\mathcal{P}(t,\vec{\eta})$ from $\vec{D}_\eta$, leading to
\begin{equation}
\mathcal{P}(t,\vec{\eta}) \sim \sum_{c=1}^{C} \pi_c \mathcal{N}(\vec{\mu}_c,\vec{\Sigma}_c),
\label{equ:gmm}
\end{equation}
where 
$\pi_c$ denotes prior probability of the $c$-th Gaussian component whose mean and covariance are, respectively, 
$\vec{\mu}_c=\left[\begin{matrix}
\vec{\mu}_{t,c} \\ \vec{\mu}_{\eta,c}
\end{matrix}\right]$ and
$\vec{\Sigma}_c=\left[\begin{matrix}
\vec{\Sigma}_{tt,c} & \vec{\Sigma}_{t\eta,c} \\ \vec{\Sigma}_{\eta t,c} & \vec{\Sigma}_{\eta \eta,c}
\end{matrix}\right]$\footnote{In order to keep notations consistent, we still use notations $\vec{\mu}_{t,c}$ and $\vec{\Sigma}_{tt,c}$ to represent scalars. }. 
Then, Gaussian mixture regression (GMR) \cite{Calinon2016, Cohn} is employed to retrieve the conditional probability distribution, i.e.,
\begin{equation}
\mathcal{P}(\vec{\eta}|t)=\sum_{c=1}^{C}h_c(t) \mathcal{N}(\bar{\vec{\mu}}_c(t),\bar{\vec{\Sigma}}_c)
\label{equ:mix:pred}
\end{equation}
with
\begin{equation}
\quad h_c(t)\!\!=\!\!\frac{\pi_c \mathcal{N}(t|\vec{\mu}_{t,c},\vec{\Sigma}_{tt,c})}{\sum_{i=1}^{C}\pi_i \mathcal{N}(t|\vec{\mu}_{t,i},\vec{\Sigma}_{tt,i})},
\end{equation} 
\begin{equation}
\bar{\vec{\mu}}_c(t)\!\!=\!\!\vec{\mu}_{\eta,c}+ \vec{\Sigma}_{\eta t,c} \vec{\Sigma}_{t t,c}^{-1} (t-\vec{\mu}_{t,c})
\end{equation}
\begin{equation}
\mathrm{and} \quad \bar{\vec{\Sigma}}_c=\vec{\Sigma}_{\eta \eta,c}- \vec{\Sigma}_{\eta t,c} \vec{\Sigma}_{t t,c}^{-1} \vec{\Sigma}_{t \eta,c}.
\end{equation} 
With the properties of multivariate Gaussian distributions, we can estimate $\mathbb{E}(\vec{\eta}|t)$ and $\mathbb{D}(\vec{\eta}|t)$ from (\ref{equ:mix:pred}), i.e., \cite{Calinon2016}
\begin{equation}
\begin{aligned}
\hat{\vec{\mu}}_t&=\mathbb{E}(\vec{\eta}|t)={\sum_{c=1}^{C}}h_c(t) \bar{\vec{\mu}}_c(t),\\
\hat{\vec{\Sigma}}_t&=\mathbb{D}(\vec{\eta}|t)=\mathbb{E}(\vec{\eta}\vec{\eta}^{\trsp}|t)-\mathbb{E}(\vec{\eta}|t)\mathbb{E}^{\trsp}(\vec{\eta}|t)\\
&=\sum_{c=1}^{C}h_c(t)\left (\bar{\vec{\mu}}_c(t) \bar{\vec{\mu}}_c^{\trsp}(t)+\bar{\vec{\Sigma}}_c \right)-\hat{\vec{\mu}}_t \hat{\vec{\mu}}^{\trsp}_t.
\end{aligned}
\end{equation}
Furthermore, we use $\mathcal{N}(\hat{\vec{\mu}}_t,\hat{\vec{\Sigma}}_t)$ to approximate (\ref{equ:mix:pred}), i.e.,
\begin{equation}
\mathcal{P}(\vec{\eta}|t)\approx\mathcal{N}(\hat{\vec{\mu}}_t,\hat{\vec{\Sigma}}_t).
\label{equ:gmr}
\end{equation}
Please refer to \cite{Calinon2016,Huang2017,Cohn} for more details. 
Therefore, for a given time sequence\footnote{The size of time sequence is not necessarily the same as that of demonstrations.} $\{t_n\}_{n=1}^{N}$ that spans the input space, we can obtain its corresponding trajectory $\{\hat{\vec{\eta}}_n\}_{n=1}^{N}$ with $\hat{\vec{\eta}}_n|t_n \sim \mathcal{N}(\hat{\vec{\mu}}_n,\hat{\vec{\Sigma}}_n)$, yielding a probabilistic reference trajectory $\vec{D}_r=\{t_n,\hat{\vec{\mu}}_n,\hat{\vec{\Sigma}}_n\}_{n=1}^{N}$. Here, we can view $\vec{D}_r$ as a representative of $\vec{D}_{\eta}$ since it encapsulates the distribution of trajectories in $\vec{D}_{\eta}$ in terms of mean and covariance. Therefore, we exploit $\vec{D}_r$ instead of the original demonstrations $\vec{D}_{\eta}$ in the next subsection.

\subsection{Learning quaternions using a kernelized approach \label{subsec:kmp}}

As a recently developed framework, KMP \cite{Huang2017} exhibits several advantages over state-of-the-art approaches:
\begin{enumerate}
	\item[(\emph{i})] In comparison with DMP \cite{Ijspeert} and TP-GMM \cite{Calinon2016}, that focus on target adaptations, KMP is capable of adapting trajectories towards arbitrary desired points (e.g., start/via/end- points). 
	\item[(\emph{ii})] Unlike DMP and ProMP \cite{Paraschos} that rely on explicit definition of basis functions, KMP employs the kernel trick to alleviate the definition of basis functions and thus allows for convenient extensions to the learning and adaptation of demonstrations consisting of high-dimensional inputs.
	\item[(\emph{iii})]In comparison to DMP and ProMP, KMP can learn complex non-linearity underlying demonstrations with fewer open parameters owing to the kernelized form.
\end{enumerate}
Note that kernel approaches generally have some limitations \cite{Bishop}, such as the storage of experience data 
and the increasing computation complexity with the size of training data. However, our approach learns the probabilistic reference trajectory instead of the raw demonstration data, thus the corresponding storage load and computation cost are alleviated.

We follow the treatment in KMP to learn the probabilistic reference trajectory  $\vec{D}_r$.
Formally, let us first write $\vec{\eta}$ in a parameterized way\footnote{Similar parametric strategies were used in DMP \cite{Ijspeert} and ProMP \cite{Paraschos}.}, i.e.,
\begin{equation}
\vec{\eta}(t)=\left[\begin{array}{c}
\vec{\zeta}(t) \\ \hdashline \dot{\vec{\zeta}}(t)
\end{array}\right]
=
\underbrace{\left[\begin{array}{cccccc}
\vec{\phi}^{\trsp}(t) & \vec{0}  & \vec{0} \\ \vec{0} & \vec{\phi}^{\trsp}(t) & \vec{0} \\ \vec{0}  & \vec{0} & \vec{\phi}^{\trsp}(t) \\ \hdashline
\dot{\vec{\phi}}^{\trsp}(t) & \vec{0}  & \vec{0} \\ \vec{0} & \dot{\vec{\phi}}^{\trsp}(t) & \vec{0} \\ \vec{0}  & \vec{0} & \dot{\vec{\phi}}^{\trsp}(t) 
\end{array}\right]}_{\vec{\Theta}^{\trsp}(t)}\vec{w},
\label{equ:para:traj}
\end{equation}
where $\vec{\phi}(t)\in \mathbb{R}^{B}$ represents a $B$-dimensional basis function vector. 
In order to learn 
$\vec{D}_r$, we consider the problem of maximizing the posterior probability
\begin{equation}
J(\vec{w})=\prod_{n=1}^{N}\mathcal{P}(\vec{\Theta}^{\trsp}(t_n)\vec{w}|\hat{\vec{\mu}}_n,\hat{\vec{\Sigma}}_n),
\label{equ:posterior}
\end{equation}
whose optimal solution $\vec{w}^{*}$ 
can be computed as
\begin{equation}
\begin{aligned}
\vec{w}^{*}\!\!=\!\mathrm{\argmin}_{\vec{w}} \!\!\sum_{n=1}^{N}\!\! \left(\vec{\Theta}^{\trsp}(t_n) \vec{w} \!-\! \hat{\vec{\mu}}_n \right)^{\trsp}   \!\!(\hat{\vec{\Sigma}}_n)^{-1} \!\!  \left(\vec{\Theta}^{\trsp}(t_n) \vec{w} \!-\! \hat{\vec{\mu}}_n \right) \\
+\lambda \vec{w}^{\trsp}\vec{w},
\end{aligned}
\label{equ:post}
\end{equation} 
where the objective to be minimized can be viewed as the sum of covariance-weighted squared errors\footnote{Similar variance-weighted scheme has also been exploited in trajectory-GMM \cite{Calinon2016}, motion similarity estimation \cite{Muhlig} and optimal control \cite{Medina}.}. Note that a regularization term $\lambda \vec{w}^{\trsp}\vec{w}$ with $\lambda>0$ is introduced in (\ref{equ:post}) so as to mitigate the over-fitting.

Similarly to the derivations of kernel ridge regression \cite{Saunders,Murphy,Kober}, the optimal solution $\vec{w}^{*}$ of (\ref{equ:post}) can be computed. Thus, for an inquiry point $t^{*}$, its corresponding output  $\vec{\eta}(t^{*})$ can be predicted as 
\begin{equation}
\begin{aligned}
\vec{\eta}(t^{*}) 
=\vec{\Theta}^{\trsp}(t^{*})\vec{w}^{*}=\vec{\Theta}^{\trsp}(t^{*})\vec{\Phi} ( \vec{\Phi}^{\trsp} \vec{\Phi} +\lambda \vec{\Sigma} )^{-1}  {\vec{\mu}}
\end{aligned}
\label{equ:pred}
\end{equation}
where 
\begin{equation}
\begin{aligned}
\vec{\Phi}&=[\vec{\Theta}(t_1) \, \vec{\Theta}(t_2) \, \cdots \, \vec{\Theta}(t_N)],\\
\vec{\Sigma}&=blockdiag(\hat{\vec{\Sigma}}_1,\hat{\vec{\Sigma}}_2, \ldots, \hat{\vec{\Sigma}}_N), \\ 
{\vec{\mu}}&=[ 
\hat{\vec{\mu}}_1^{\trsp} \, \hat{\vec{\mu}}_2^{\trsp} \, \cdots \, \hat{\vec{\mu}}_N^{\trsp}]^{\trsp}.
\end{aligned}
\label{equ:mat}
\end{equation}
Furthermore, (\ref{equ:pred}) can be kernelized as
\begin{equation}
\vec{\eta}(t^{*}) 
=\vec{k}^{*}( \vec{K} +\lambda \vec{\Sigma} )^{-1}  {\vec{\mu}}
\label{equ:kernel:pred}
\end{equation}
with $\vec{k}^{*}_{[i]}=\vec{k}(t^{*},t_i)$ and $\vec{K}_{[i,j]}=\vec{k}(t_i,t_j)$, $i\in\{1,2,\ldots,N\}, j\in\{1,2,\ldots,N\}$, where
$\vec{k}(\cdot,\cdot)$ is defined by
\begin{equation}
\begin{aligned}
\vec{k}(t_i,t_j)
=\vec{\Theta}^{\trsp}(t_i)\vec{\Theta}(t_j)
=\left[\begin{matrix}
k_{t,t}(t_i,t_j) \vec{I}_3 &\! \!k_{t,d}(t_i,t_j)  \vec{I}_3\\
k_{d,t}(t_i,t_j)\vec{I}_3 & \! \!k_{d,d}(t_i,t_j) \vec{I}_3
\end{matrix}\right]
\end{aligned}
\label{equ:kernel}
\end{equation}
with\footnote{Note that $\dot{\vec{\phi}}(t)$ is approximated by $\dot{\vec{\phi}}(t)\approx \frac{{\vec{\phi}}(t+\delta)-{\vec{\phi}}(t)}{\delta}$ in order to facilitate the following kernelized operations.}
\begin{equation*}
\begin{aligned}
k_{t,t}(t_i,t_j)&\!\!=\!\!k(t_i,t_j),\\
k_{t,d}(t_i,t_j)&\!\!=\!\!\frac{k(t_i,t_j+\delta) -k(t_i,t_j)}{\delta},\\
k_{d,t}(t_i,t_j)&\!\!=\!\!\frac{k(t_i+\delta,t_j)-k(t_i,t_j)}{\delta},\\
k_{d,d}(t_i,t_j)&\!\!=\!\!\frac{k(t_i\!\!+\!\!\delta, t_j\!\!+\!\!\delta)\!\!-\!\!k(t_i\!\!+\!\!\delta,t_j)\!\!-\!\!k(t_i ,t_j\!\!+\!\delta)\!\!+\!k(t_i,t_j)}{\delta^2} ,
\end{aligned}
\end{equation*}  
where $\delta>0$ is a small constant
and $k(t_i,t_j)=\vec{\phi}(t_i)^{\trsp}\vec{\phi}(t_j)$ represents the kernel function. 

By observing  (\ref{equ:para:traj}), we can find that both $\vec{\phi}(t)$ and $\dot{\vec{\phi}}(t)$ are used. If we would have used $\vec{\phi}(t)$ to parameterize $\vec{\zeta}(t)$ and $ \dot{\vec{\zeta}}(t)$ independently of each other, i.e.,  $\vec{\eta}(t)=\left[\begin{array}{c}
\vec{\zeta}(t) \\  \dot{\vec{\zeta}}(t)
\end{array}\right]=diag\bigl(\vec{\phi}^{\trsp}(t),\vec{\phi}^{\trsp}(t),\ldots,\vec{\phi}^{\trsp}(t)\bigr)\left[\begin{array}{c}
\vec{w}_1 \\  \vec{w}_2
\end{array}\right]$ with $\vec{w}_1\in\mathbb{R}^{3B}$ and $\vec{w}_2\in\mathbb{R}^{3B}$, a simpler kernel $\vec{k}(t_i,t_j)=k(t_i,t_j)\vec{I}_6$ can be obtained in (\ref{equ:kernel}). However, this treatment would ignore the derivative relationship between $\vec{\zeta}(t)$ and $\dot{\vec{\zeta}}(t)$. Consequently, in predictions (\ref{equ:pred}) and (\ref{equ:kernel:pred}), the derivative relationship between $\vec{\zeta}(t^{*})$ and $\dot{\vec{\zeta}}(t^{*})$ could not be guaranteed.

Once we have determined $\vec{\eta}(t^{*})$ at a query point $t^{*}$ via (\ref{equ:kernel:pred}), we can use its component $\vec{\zeta}(t^{*})$ to recover the corresponding quaternion $\vec{q}(t^{*})$. Specifically,
$\vec{q}(t^{*})$ is determined by
\begin{equation}
\vec{q}(t^{*})=\exp(\vec{\zeta}(t^{*})) * \vec{q}_a,
\label{equ:recover}
\end{equation}
where the function $\exp(\cdot): \mathbb{R}^{3}\mapsto \mathbb{S}^{3}$ is \cite{Ude,Abu}
\begin{equation}
\exp(\vec{\zeta})=
\left\{ 
\begin{aligned}
&
\begin{bmatrix}
\cos(||\vec{\zeta}||) \\ \sin(||\vec{\zeta}||)\frac{\vec{\zeta}}{||\vec{\zeta}||}
\end{bmatrix}
, \;\; \vec{\zeta}\neq \vec{0} \\
&[1\;\;0\;\; 0 \;\;0]^{\trsp}, \quad\quad\quad\quad \quad otherwise.
\end{aligned}  
\right.
\label{equ:quat:exp}
\end{equation}

It should be noted that the singularity issue exists in $\log(\cdot)$, thus an assumption is imposed throughout this paper:
	
\textbf{{Assumption 2} \cite{Ude}} \emph{The input domain of the mapping $\log(\cdot)$ is restricted to $ \mathbb{S}^3$ except for $[-1 \ 0 \ 0 \ 0]^{\trsp}$, while the input domain of the mapping $\exp(\vec{\zeta})$ should fulfill the constraint $||\vec{\zeta}||<\pi$. }

Under this assumption, both $\log(\cdot)$ and $\exp(\cdot)$ are bijective, and $\exp(\cdot)$ can be viewed as the inverse function of $\log(\cdot)$, leading to $\exp\bigl(\log(\vec{q}*\bar{\vec{q}}_{a})\bigr)*\vec{q}_{a}=\vec{q}$. Please refer to \cite{Ude} for the discussion. Note that when we choose the auxiliary quaternion $\vec{q}_a$ in (\ref{equ:transfer}), it must obey the Assumption 2.

\section{Adaptation of Quaternion Trajectories \label{sec:ada}} 

While the approach in Section~\ref{subsec:kmp} is limited to orientation reproduction, we now consider the problem of adapting the reference trajectory in terms of desired quaternions and angular velocities. To do so, we propose to transform desired orientation states into Euclidean space (Section~\ref{subsec:trans:des}), and subsequently we reformulate the kernelized learning approach to incorporate these transformed desired points (Section~\ref{subsec:ada:kmp}). Finally, the adapted trajectory in Euclidean space is used to retrieve its corresponding adapted quaternion trajectory.

\subsection{Transform desired quaternion states \label{subsec:trans:des}}
Let us denote $H$ desired quaternion states as $\tilde{\vec{D}}_q=\{\tilde{t}_h,\tilde{\vec{q}}_h,\tilde{\vec{\omega}}_h\}_{h=1}^{H}$, where $\tilde{\vec{q}}_h \in \mathbb{S}^3$ and $\tilde{\vec{\omega}}_h \in \mathbb{R}^3$ represent desired quaternion and angular velocity at time $\tilde{t}_h$, respectively. 
Similarly to (\ref{equ:transfer}), 
the desired quaternion $\tilde{\vec{q}}_h$ can be transformed as
\begin{equation}
\tilde{\vec{\zeta}}_h=\log( \tilde{\vec{q}}_h * \bar{\vec{q}}_a).
\label{equ:des:quat:transfer}
\end{equation}
In order to incorporate the desired angular velocity $\tilde{\vec{\omega}}_h$, we resort to the relationship between derivatives of quaternions and angular velocities, i.e., \cite{Ude,Abu}
\begin{equation}
\dot{\vec{q}}=\frac{1}{2}
\begin{bmatrix}0 \\ \vec{\omega} \end{bmatrix}
*\vec{q} \Rightarrow {\vec{q}}(t+\delta_t)=\exp\left(\frac{\vec{\omega}}{2}\delta_t\right)*\vec{q}(t),
\label{equ:quat:angular} 
\end{equation}
where $\delta_t>0$ denotes a small constant. 
By using (\ref{equ:quat:angular}), we can compute the desired quaternion at time $\tilde{t}_h+\delta_t$ as
\begin{equation}
	\tilde{\vec{q}}(\tilde{t}_h+\delta_t)=\exp\left(\frac{\tilde{\vec{\omega}}_h}{2}\delta_t\right)*\tilde{\vec{q}}_h,
\label{equ:des:quat:dt}
\end{equation}
which is subsequently transformed into Euclidean space via (\ref{equ:transfer}), resulting in
\begin{equation}
\begin{aligned}
\tilde{\vec{\zeta}}(\tilde{t}_h+\delta_t)=\log( \tilde{\vec{q}}(\tilde{t}_h+\delta_t) * \bar{\vec{q}}_a).
\label{equ:des:quat:transfer:dt}
\end{aligned}
\end{equation}
Thus, 
we can approximate the derivative of $\tilde{\vec{\zeta}}_h$ as
\begin{equation}
\begin{aligned}
\dot{\tilde{\vec{\zeta}}}_h &\approx \frac{\tilde{\vec{\zeta}}(\tilde{t}_h+\delta_t)-\tilde{\vec{\zeta}}_h}{\delta_t}\\
&=\frac{\log \left( \left(\exp(\frac{\tilde{\vec{\omega}}_h}{2}\delta_t)*\tilde{\vec{q}}_h\right) * \bar{\vec{q}}_a \right) \!\!-\! \log( \tilde{\vec{q}}_h * \bar{\vec{q}}_a)}{\delta_t}.
\end{aligned}
\label{equ:des:der:tranfer}
\end{equation}
Now, 
$\tilde{\vec{D}}_q$ can be transformed into $\tilde{\vec{D}}_{\zeta}=\{\tilde{t}_h,\tilde{\vec{\zeta}}_h,\dot{\tilde{\vec{\zeta}}}_h\}_{h=1}^{H}$ via (\ref{equ:des:quat:transfer}) and (\ref{equ:des:der:tranfer}), which can be further rewritten in a compact way as $\tilde{\vec{D}}_{\eta}=\{\tilde{t}_h,\tilde{\vec{\eta}}_h\}_{h=1}^{H}$ with $\tilde{\vec{\eta}}_h=[\tilde{\vec{\zeta}}_h^{\trsp}\,\dot{\tilde{\vec{\zeta}}}_h^{\trsp}]^{\trsp} \in \mathbb{R}^{6}$. In addition, we can design a covariance
 $\tilde{\vec{\Sigma}}_h \in \mathbb{R}^{6\times 6}$ for each desired point $\tilde{\vec{\eta}}_h$ to control the precision of adaptations. 
Thus, we can obtain an additional probabilistic reference trajectory  $\tilde{\vec{D}}_r=\{\tilde{t}_h,\tilde{\vec{\eta}}_h,\tilde{\vec{\Sigma}}_h\}_{h=1}^{H}$ to indicate the transformed desired quaternion states.

\subsection{Adapting quaternion trajectories towards desired points \label{subsec:ada:kmp}}

Formally, the adaptation problem can be addressed by incorporating $\tilde{\vec{D}}_{r}$ into (\ref{equ:post}), i.e.,
\begin{equation}
\begin{aligned}
&\vec{w}^{*}=\mathrm{\argmin}_{\vec{w}} \!\!\underbrace{\sum_{n=1}^{N}\!\! \left(\vec{\Theta}^{\trsp}(t_n) \vec{w} \!-\! \hat{\vec{\mu}}_n \right)^{\trsp}   \!\!(\hat{\vec{\Sigma}}_n)^{-1} \!\!  \left(\vec{\Theta}^{\trsp}(t_n) \vec{w} \!-\! \hat{\vec{\mu}}_n \right)}_{imitation}\\ 
&+ \underbrace{\sum_{h=1}^{H}\!\! \left(\vec{\Theta}^{\trsp}(\tilde{t}_h) \vec{w} \!-\! \tilde{\vec{\eta}}_h \right)^{\trsp}   \!\!(\tilde{\vec{\Sigma}}_h)^{-1} \!\!  \left(\vec{\Theta}^{\trsp}(\tilde{t}_h) \vec{w} \!-\!  \tilde{\vec{\eta}}_h \right)}_{adaptations}
+\underbrace{\lambda \vec{w}^{\trsp}\vec{w}}_{regularizer},
\end{aligned}
\label{equ:post:new}
\end{equation}  
whose compact representation is
\begin{equation}
\begin{aligned}
\vec{w}^{*}=\mathrm{\argmin}_{\vec{w}} \sum_{l=1}^{N+H} &\biggl( \left(\vec{\Theta}^{\trsp}({t}_l^{U}) \vec{w} - {\vec{\mu}}_l^{U} \right)^{\trsp}   ({\vec{\Sigma}}_l^{U})^{-1}  \\ &\left(\vec{\Theta}^{\trsp}({t}_l^{U}) \vec{w} - {\vec{\mu}}_l^{U} \right) \biggr) 
+\lambda \vec{w}^{\trsp}\vec{w},
\end{aligned}
\label{equ:post:compact}
\end{equation}    
with 
 \begin{equation}
 \left\{\!\!
 \begin{aligned}
 &{t}_l^{U}=t_l,{\vec{\mu}}_l^{U}=\hat{\vec{\mu}}_l,{\vec{\Sigma}}_l^{U}=\hat{\vec{\Sigma}}_l, \quad \quad \quad \quad 1 \leq l \leq N \\ 
 &{t}_l^{U}=\tilde{t}_{l-N},{\vec{\mu}}_l^{U}\!=\tilde{\vec{\eta}}_{l-N},{\vec{\Sigma}}_l^{U}\!\!=\tilde{\vec{\Sigma}}_{l-N}, N\!+\!1 \leq l \leq N\!\!+\!\!H.
 \end{aligned}
 \right.
 \label{equ:ref:update}
 \end{equation}
It can be observed that the new objective (\ref{equ:post:compact}) shares the same form with (\ref{equ:post}), except that the reference trajectory ${\vec{D}}_r^{U}=\{{t}_l^{U},{\vec{\mu}}_l^{U},{\vec{\Sigma}}_l^{U}\}_{l=1}^{N+H}$ in (\ref{equ:post:compact}) is longer than that in (\ref{equ:post}), thus the solution of (\ref{equ:post:compact}) can be determined in a similar way. Finally, $\vec{\eta}(t)=[\vec{\zeta}^{\trsp}(t) \, \dot{\vec{\zeta}}^{\trsp}(t)]^{\trsp}$ can be computed via (\ref{equ:kernel:pred}) and, subsequently, $\vec{q}(t)$ is recovered from (\ref{equ:recover}) by using $\vec{\zeta}(t)$. In this case, $\vec{q}(t)$ is capable of passing through the desired quaternions $\tilde{\vec{q}}_h$ with desired angular velocities $\tilde{\vec{\omega}}_h$ at time $\tilde{t}_h$, provided that $\tilde{\vec{\Sigma}}_h$ is small enough\footnote{In (\ref{equ:post:new}) both the `imitation' and `adaptation'  have impacts on $\vec{w}^{*}$ and both terms rely on their own covariance matrices. Thus,
if one needs precise adaptation, $\tilde{\vec{\Sigma}}_h$ should be set far smaller (e.g., at least 2-3 orders of magnitude) than the variance of the reference trajectory.}.

\section{Quaternion Adaptations with Angular Acceleration/Jerk Constraints\label{sec:quat:const}}

It is well known that robot trajectories should be smooth in order to facilitate the design of controllers as well as the execution of motor commands \cite{Koc,Ratliff}. For instance, in a striking task that needs fast striking motions, extremely high accelerations or jerks may degrade the final striking performance, given the physical limits of motors. It is possible to formulate this constraint as an optimization problem and search for the optimal trajectory via an iterative scheme, as done in \cite{Koc}. 
In this section, we consider the problem of learning and adapting quaternion trajectories while taking into account angular acceleration or jerk constraints. Specifically, we aim to provide an \emph{analytical} solution to the issue.

Formally, we consider the angular acceleration or jerk constraints as minimizing $\sum_{n=1}^{N}||\dot{\vec{\omega}}(t_n)||^2$ or $\sum_{n=1}^{N}||\ddot{\vec{\omega}}(t_n)||^2$. Note that the aforementioned imitation learning problem (\ref{equ:post}) is built on the trajectory $\vec{\zeta}(t)$, 
therefore we need to find the relationship between $\vec{\omega}(t)$ and $\vec{\zeta}(t)$. Here, we provide two main results:

\textbf{{Theorem 1}} \emph{Given the definition $\vec{\zeta}(t)=\log(\vec{q}(t)*\bar{\vec{q}}_{a}) $, if we let
$\vec{q}(t_1)=\vec{q}_a$, then the optimal quaternion trajectory $\{\vec{q}(t_n)\}_{t=1}^{N+2}$ of minimizing $\sum_{n=1}^{N}||\ddot{\vec{\zeta}}(t_n)||^2$ corresponds to the optimal solution of minimizing $\sum_{n=1}^{N}||\dot{\vec{\omega}}(t_n)||^2$}.
\begin{proof}
The minimization of $\sum_{n=1}^{N}||\ddot{\vec{\zeta}}(t_n)||^2$ implies that \begin{equation*}
\dot{\vec{\zeta}}(t_1)=\dot{\vec{\zeta}}(t_2)=\cdots=\dot{\vec{\zeta}}(t_N)=\dot{\vec{\zeta}}(t_{N+1}).
\end{equation*}
Consequently, we can write
$\vec{\zeta}(t_{n+1})=\vec{\zeta}(t_n)+\Delta$ with $\Delta\in \mathbb{R}^3$ being a constant. Given $\vec{q}(t_1)=\vec{q}_a$, we have $\vec{\zeta}(t_1)=\vec{0}$ and 
\begin{equation*}
\vec{\zeta}(t_n)=(n-1)\Delta, n=1,2,\ldots,N+2.
\end{equation*}
Using the definition of $\vec{\zeta}(t)$,
we can obtain the optimal quaternion trajectory $\{\vec{q}(t_n)\}_{t=1}^{N+2}$ of minimizing $\sum_{n=1}^{N}||\ddot{\vec{\zeta}}(t_n)||^2$, i.e.,  
\begin{equation*}
\vec{q}(t_n)=\exp\bigl((n-1)\Delta\bigr)* \vec{q}_a.
\end{equation*} 
Now, we use the optimal quaternion trajectory to calculate the corresponding $\{\vec{\omega}(t_n)\}_{n=1}^{N+1}$. Specifically, we have\footnote{The following results are used in Theorems 1 and 2:
		\begin{enumerate}
			\item [(\emph{i})] If $\Delta \neq \vec{0}$ and $i \neq j$, 
	\begin{equation*}\begin{aligned}
&\exp(i\Delta) * \overline{\exp(j\Delta)}
    =\begin{bmatrix}
	cos (||i\Delta||) \\ sin (||i\Delta||)\frac{\Delta}{||\Delta||}
	\end{bmatrix}* \begin{bmatrix}
	cos (||j\Delta||) \\ -sin (||j\Delta||)\frac{\Delta}{||\Delta||}
	\end{bmatrix}\\
&	=\begin{bmatrix}
	cos (||i\Delta||-||j\Delta||) \\ sin (||i\Delta||-||j\Delta||)\frac{\Delta}{||\Delta||}
	\end{bmatrix}
	=\begin{bmatrix}
	cos ((i-j)||\Delta||) \\ sin ((i-j)||\Delta||)\frac{\Delta}{||\Delta||}
	\end{bmatrix}\\
&	=\begin{bmatrix}
	cos (|i-j|||\Delta||) \\ sin (|i-j|||\Delta||)\frac{(i-j)\Delta}{|i-j|||\Delta||}
	\end{bmatrix}
	=\exp((i-j)\Delta).
	\end{aligned}\end{equation*}
	\item [(\emph{ii})] If $\Delta = \vec{0}$ or $i = j$, 
 \begin{equation*}
\exp(i\Delta) * \overline{\exp(j\Delta)}\!=\begin{bmatrix}
1 \\  \vec{0}
\end{bmatrix}\!=\!\exp((i-j)\Delta).	 
\end{equation*} 	
		\end{enumerate}
}
\begin{equation*}
\begin{aligned}
\vec{\omega}(t_n)&=\frac{2}{\delta_t}\log\bigl(\vec{q}(t_{n+1})*\overline{\vec{q}(t_{n})}\bigr)\\
&=\frac{2}{\delta_t}\log\bigl(\exp(n\Delta)*\overline{\exp((n-1)\Delta)}\bigr)\\
&=\frac{2}{\delta_t} \log(\exp(\Delta))=\frac{2}{\delta_t}\Delta
\end{aligned}
\end{equation*}
with $\delta_t$ being the time interval between $\vec{q}(t_{n})$ and $\vec{q}(t_{n+1})$,
which implies
\begin{equation*}
\vec{\omega}(t_1)=\vec{\omega}(t_2)=\cdots=\vec{\omega}(t_N)=\vec{\omega}(t_{N+1}).
\end{equation*}
Thus, we have $\dot{\vec{\omega}}(t_n)=\vec{0}, n=1,2,\ldots,N$, which corresponds to the optimal solution of minimizing $\sum_{n=1}^{N}||\dot{\vec{\omega}}(t_n)||^2$.
\end{proof}

\textbf{{Theorem 2}} \emph{Given the definition $\vec{\zeta}(t)=\log(\vec{q}(t)*\bar{\vec{q}}_{a}) $, if we let $\vec{q}(t_1)=\vec{q}_a$ and $\vec{\omega}(t_1)=\vec{0}$, then the optimal quaternion trajectory $\{\vec{q}(t_n)\}_{t=1}^{N+3}$ of minimizing $\sum_{n=1}^{N}||\dddot{\vec{\zeta}}(t_n)||^2$ corresponds to the optimal solution of minimizing $\sum_{n=1}^{N}||\ddot{\vec{\omega}}(t_n)||^2$}.
\begin{proof}
Minimizing $\sum_{n=1}^{N}||\dddot{\vec{\zeta}}(t_n)||^2$ corresponds to 
\begin{equation*}
	\ddot{\vec{\zeta}}(t_1)=\ddot{\vec{\zeta}}(t_2)=\cdots=\ddot{\vec{\zeta}}(t_N)=\ddot{\vec{\zeta}}(t_{N+1}).
\end{equation*} 
	Then, we have
	$\dot{\vec{\zeta}}(t_{n+1})=\dot{\vec{\zeta}}(t_n)+\Delta$, where $\Delta\in \mathbb{R}^3$ is a constant. With the approximation $\dot{\vec{\zeta}}(t)=\frac{\vec{\zeta}(t+\delta_t)-\vec{\zeta}(t)}{\delta_t}$, we have $\vec{\zeta}(t_{n+2})= 2\vec{\zeta}(t_{n+1})-\vec{\zeta}(t_{n})+\delta_t \Delta$. 
	Note that we assume $\vec{q}(t_1)=\vec{q}_a$ and $\vec{\omega}(t_1)=0$. Hence, $\vec{q}(t_1)=\vec{q}(t_2)=\vec{q}_a$ and $\vec{\zeta}(1)=\vec{\zeta}(2)=0$. It can be further seen that 
	\begin{equation*}
	\vec{\zeta}(t_{n})=\frac{(n-1)(n-2)}{2}\delta_t\Delta, n=1,2,\ldots,N+3.
	\end{equation*}
	Using the definition of $\vec{\zeta}(t)$, we have the optimal quaternion trajectory $\{\vec{q}(t_n)\}_{t=1}^{N+3}$ of minimizing $\sum_{n=1}^{N}||\dddot{\vec{\zeta}}(t_n)||^2$, i.e.,  
	\begin{equation*}
\vec{q}(t_n)= \exp\left( \frac{(n-1)(n-2)}{2}\delta_t\Delta  \right) * \vec{q}_a.
	\end{equation*}
	The corresponding angular velocity $\{\vec{\omega}(t_n)\}_{n=1}^{N+2}$ of using the optimal quaternion trajectory is
	\begin{equation*}
	\begin{aligned}
&\vec{\omega}(t_n)=\frac{2}{\delta_t}\log \left(\vec{q}(t_{n+1})*\overline{\vec{q}(t_{n})}\right)\\ &=\frac{2}{\delta_t}\!\log\!\left(\!\exp\!\!\left(\!\! \frac{n(n-1)}{2}\delta_t\Delta  \!\!\right) * \overline{\exp\left(\!\! \frac{(n-1)(n-2)}{2}\delta_t\Delta  \!\!\right)}\right)\\
	&=\frac{2}{\delta_t}\log\left( \exp \bigl( (n-1)\delta_t\Delta\bigr) \right)
	=2(n-1)\Delta,
		\end{aligned}
	\end{equation*} 
	which implies that $\vec{\omega}(t_{n+1})-\vec{\omega}(t_{n})=2\Delta$. 
	Thus,  
	\begin{equation*}
	\dot{\vec{\omega}}(t_1)=\dot{\vec{\omega}}(t_2)=\cdots=\dot{\vec{\omega}}(t_N)=\dot{\vec{\omega}}(t_{N+1}),
	\end{equation*} 
	leading to $\ddot{\vec{\omega}}(t_n)=\vec{0},  n=1,2,\ldots,N$. So, we can conclude that the optimal trajectory $\{\vec{q}(t_n)\}_{n=1}^{N+3}$ that minimizes $\sum_{n=1}^{N}||\dddot{\vec{\zeta}}(t_n)||^2$ yields the optimal solution $\{\vec{\omega}(t_n)\}_{n=1}^{N+2}$ of minimizing $\sum_{n=1}^{N}||\ddot{\vec{\omega}}(t_n)||^2$.
\end{proof}

With Theorems 1 and 2, the problem of learning quaternions with angular acceleration or jerk constraints can be approximately tackled by incorporating $\sum_{n=1}^{N}||\ddot{\vec{\zeta}}(t_n)||^2$ or $\sum_{n=1}^{N}||\dddot{\vec{\zeta}}(t_n)||^2$ into the objective (\ref{equ:post}). For brevity, we take the angular acceleration constraints as an example, while the case of angular jerk constraints can be treated in a similar way. Following the parameterization form in (\ref{equ:para:traj}), we have
\begin{equation}
\ddot{\vec{\zeta}}(t)=
\underbrace{\left[\!\begin{array}{ccc}
\ddot{\vec{\phi}}^{\trsp}(t) & \vec{0}  & \vec{0} \\ \vec{0} & \ddot{\vec{\phi}}^{\trsp}(t) & \vec{0} \\ \vec{0}  & \vec{0} & \ddot{\vec{\phi}}^{\trsp}(t) 
\end{array}\!\right]}_{\vec{\varphi}^{\trsp}(t)}\vec{w}.
\label{equ:const:notation}
\end{equation} 
Thus, the problem of learning orientations with angular acceleration constraints becomes  
\begin{equation}
\begin{aligned}
\vec{w}^{*}\!\!=\!\mathrm{\argmin}_{\vec{w}} \!\!\underbrace{\sum_{n=1}^{N}\!\! \left(\vec{\Theta}^{\trsp}(t_n) \vec{w} \!-\! \hat{\vec{\mu}}_n \right)^{\trsp}   \!\!(\hat{\vec{\Sigma}}_n)^{-1} \!\!  \left(\vec{\Theta}^{\trsp}(t_n) \vec{w} \!-\! \hat{\vec{\mu}}_n \right)}_{imitation} \\
+ {\lambda}_a \underbrace{\sum_{n=1}^{N}(\vec{\varphi}^{\trsp}(t_n)\vec{w})^{\trsp}(\vec{\varphi}^{\trsp}(t_n)\vec{w})}_{angular \; acceleration\; constraints}+\underbrace{\lambda \vec{w}^{\trsp}\vec{w}}_{regularizer},
\end{aligned}
\label{equ:post:const}
\end{equation} 
where ${\lambda}_a>0$ acts as a trade-off regulator between orientation learning and angular acceleration minimization.

Let us re-arrange (\ref{equ:post:const}) into a compact form, resulting in
\begin{equation}
\begin{aligned}
\vec{w}^{*}\!\!=\!\mathrm{\argmin}_{\vec{w}} \!\!\sum_{n=1}^{N}\!\! \left(\vec{\Omega}^{\trsp}(t_n) \vec{w} \!-\! \bar{\vec{\mu}}_n \right)^{\trsp}   \!\!(\bar{\vec{\Sigma}}_n)^{-1} \!\!  \left(\vec{\Omega}^{\trsp}(t_n) \vec{w} \!-\! \bar{\vec{\mu}}_n \right) \\
+\lambda \vec{w}^{\trsp}\vec{w},
\end{aligned}
\label{equ:quat:comp:jerk}
\end{equation} 
where
\begin{equation}
\vec{\Omega}(t_n)\!\!=\!\!\left[
		\vec{\Theta}(t_n) \ {\vec{\varphi}}(t_n)\right], 
\bar{\vec{\mu}}_n\!\!=\!\!\left[\begin{matrix}\hat{\vec{\mu}}_n \\
		\vec{0}\end{matrix}\right], 
\bar{\vec{\Sigma}}_n\!\!=\!\!\left[\begin{matrix} 
	\hat{\vec{\Sigma}}_n & \vec{0}\\
	\vec{0} & \frac{1}{\lambda_a}
	\vec{I}\end{matrix}\right].
\label{equ:notation:add:acc}
\end{equation}
It can be observed that (\ref{equ:quat:comp:jerk}) shares the same formula as (\ref{equ:post}), and hence we can follow (\ref{equ:pred})--(\ref{equ:quat:exp}) to derive a kernelized solution for quaternion reproduction with angular acceleration constraints. Note that $\vec{\Omega}(t)$ comprises the second-order derivative of $\vec{\phi}(t)$, thus a new kernel matrix \begin{equation}
\vec{k}(t_i,t_j)=\vec{\Omega}^{\trsp}(t_i)\vec{\Omega}(t_j)=
\left[\!\begin{matrix}
\vec{\Theta}^{\trsp}(t_i){\vec{\Theta}}(t_j) & \vec{\Theta}^{\trsp}(t_i){\vec{\varphi}}(t_j)\\  
\vec{\varphi}^{\trsp}(t_i){\vec{\Theta}}(t_j) & \vec{\varphi}^{\trsp}(t_i){\vec{\varphi}}(t_j)
\end{matrix}\right]
\label{equ:kernel:const}
\end{equation}
is required instead of (\ref{equ:kernel}). Please see the detailed derivations of kernelizing (\ref{equ:kernel:const}) in Appendix~\ref{app:kernel}.

Similarly, by analogy with (\ref{equ:post}) and (\ref{equ:post:new}), the adaptation issue with angular acceleration or jerk constraints can be addressed by reformulating (\ref{equ:quat:comp:jerk}) to include the desired points. It is noted that the assumptions in Theorem 1 (i.e., $\vec{q}(t_1)=\vec{q}_a$) and Theorem 2 (i.e., $\vec{q}(t_1)=\vec{q}_a$ and  $\vec{\omega}(t_1)=\vec{0}$) are trivial since they can be guaranteed by simply specifying a desired point at time $t_1$.

\section{Learning Quaternions Associated with High-Dimensional Inputs \label{sec:ori:highDim}}

While the aforementioned results focus on learning and adapting orientation trajectories associated with time input, we now consider the case of learning orientations with high-dimensional varying inputs. 
Specifically, we focus on the problem of learning demonstrations 
$\vec{D}^{0}=\{\{\vec{s}_{n,m}, \vec{\xi}_{n,m}^{0}\}_{n=1}^{N}\}_{m=1}^{M}$ consisting of inputs 
$\vec{s}_{n,m}$
and outputs\footnote{Note that 
outputs that comprise multiple Cartesian positions, quaternions, rotation matrices and joint positions can be tackled similarly.} $\vec{\xi}_{n,m}^{0}=\left[\begin{matrix}
\vec{p}_{n,m} \\ \vec{q}_{n,m}
\end{matrix}\right] $, where
$\vec{s}_{n,m} \in \mathbb{R}^\mathcal{I}$ denotes an $\mathcal{I}$-dimensional input vector and $\vec{\xi}_{n,m}^{0}$ stands for the concatenation of end-effector position $\vec{p}$ and quaternion $\vec{q}$ in Cartesian space. 
Please note that the varying high-dimensional input trajectories $\{\{\vec{s}_{n,m}\}_{n=1}^{N}\}_{m=1}^{M}$ are parts of demonstrations, which shall not be confused with external contextual variables describing the conditions under which demonstrations are recorded.

In order to illustrate the importance of learning demonstrations that consist of multiple varying inputs, we first motivate this problem in Section~\ref{subsec:quat:moti:higDim}. After that, we show the modeling of demonstrations $\vec{D}^{0}$ in Section~\ref{subsec:quat:model:higDim}, which is later exploited to derive the kernelized approach for learning and adapting quaternions associated with high-dimensional inputs (Sections~\ref{subsec:quat:learn:higDim} and \ref{subsec:quat:ada:higDim}).

\subsection{Why learning demonstrations comprising high-dimensional inputs?\label{subsec:quat:moti:higDim}}
Let us take human-robot collaboration as an example, where the robot is demanded to react properly in response to the human states (e.g., human hand positions/orientations).
In many previous work,
human and robot motions are encoded by taking time as input (e.g.,  DMP was used in \cite{Amor} and ProMP in \cite{Maeda}). Specifically, when the human trajectory is rescaled in  time, the corresponding robot trajectory with respect to time will also be modified. However, this treatment will cause a synchronization issue, since human motions in the new evaluations could be significantly different (e.g., faster/slower velocity) from the demonstrated ones. For instance, assuming in a human-robot handover task where demonstrations (including human and robot motions) lasting for $10s$ are recorded, but in the evaluation stage human hand has a pause for more than $10s$ before moving. In this case, the robot will still keep moving as its trajectory is driven by time. As a consequence, before the human hand starts to move, the robot has finished its hand-over action, 
which violates the synchronization constraints. To avoid this issue, in \cite{Ewerton} human movement duration in new evaluations is required to be the same as the one in the training demonstrations. However, as pointed out in \cite{Maeda}, this restriction on human motion duration is impractical. In order to provide a generic solution for human-robot collaboration, various strategies of phase-estimation and time-alignment are designed towards synchronizing human and robot in \cite{Amor, Maeda}. 

In contrast, we propose to consider high-dimensional varying signals (e.g., human hand positions/orientations) as inputs and predict robot motions (e.g., Cartesian positions and orientations) according to the sensed states of the human hand. We have successfully tested this solution in previous work on human-robot collaboration, namely in the collaborative hand task \cite{Huang2017} and robot-assisted painting task \cite{Joao2019,Joao2018Iros}. However, in none of those works we have considered orientation outputs, as the proposed tools were designed for Euclidean data, and only \cite{Huang2017} considers adaptation to new, unseen inputs. Note that now robot trajectory is directly driven by the state of the human hand, thus time is not explicitly involved when predicting robot actions. Recalling the above-mentioned example and following our strategy, if the human hand has a pause (i.e., inputs are unchanged), the corresponding robot trajectory (i.e., outputs) will also remain unchanged. Therefore, the main advantage of learning demonstrations consisting of multiple inputs is that additional synchronization procedures are not needed, providing a straightforward solution for accomplishing complex collaboration tasks.

\subsection{Modeling quaternions with high-dimensional inputs\label{subsec:quat:model:higDim}} 

Handling high-dimensional inputs requires a different treatment, when compared to the time-driven case described in Section~\ref{sec:model:demo}, due to the higher complexity of the input space. 
Since we are considering quaternions as outputs, we follow the procedure in Section~\ref{subsec:gmm}, transferring $\vec{D}^{0}$ into $\vec{D}^s=\{\{\vec{s}_{n,m}, \vec{\xi}_{n,m}\}_{n=1}^{N}\}_{m=1}^{M}$, where  
$\vec{\xi}_{n,m}\!\!=\!\!\left[\begin{matrix}
\vec{p}_{n,m} \\ \log(\vec{q}_{n,m}*\bar{\vec{q}}_{a})
\end{matrix}\right] \!\in\! \mathbb{R}^{6} $. Then, we model the joint probability distribution $\mathcal{P}(\vec{s},\vec{\xi})$ from $\vec{D}^s$ via GMM, i.e.,
\begin{equation}
\mathcal{P}(\vec{s},\vec{\xi}) \sim \sum_{c=1}^{C} \pi_c \mathcal{N}(\vec{\mu}_c,\vec{\Sigma}_c)
\label{equ:gmm:highD}
\end{equation}
with 
$\vec{\mu}_c=\left[\begin{matrix}
\vec{\mu}_{\vec{s},c} \\ \vec{\mu}_{\vec{\xi},c}
\end{matrix}\right]$ and
$\vec{\Sigma}_c=\left[\begin{matrix}
\vec{\Sigma}_{\vec{s}\vec{s},c} & \vec{\Sigma}_{\vec{s}\vec{\xi},c} \\ \vec{\Sigma}_{\vec{\xi}\vec{s},c} & \vec{\Sigma}_{\vec{\xi}\vec{\xi},c}
\end{matrix}\right]$.

However, unlike the generation of probabilistic reference trajectories from time-driven demonstrations (Section~\ref{subsec:gmm}), it is not straightforward to decide the input sequence $\{\vec{s}_n\}_{n=1}^{N}$ for retrieving the corresponding reference trajectory, due to the fact that ${\vec{s}}$ is high-dimensional. 

Note that a proper input sequence should span the whole input space, in order to adequately encapsulate all shown robot behaviors. This is relatively straightforward when the input is time, since the inputs of all demonstrations lie on one axis and have roughly the same duration.
When the input is high-dimensional, one intuitive solution is to use the input parts of all demonstrations, which, however, will lead to two issues: (\emph{i}) if all training inputs are used, redundancy will often arise, leading to a data-inefficient solution, where multiple data-points map to the same robot pose; (\emph{ii}) if only parts of input trajectories are exploited, one risks failing to capture important input points.

Therefore, we propose to sample inputs from the marginal probability distribution\footnote{Readers are suggested to refer to \cite{Hershey} for GMM sampling.} $\mathcal{P}(\vec{s})$. Specifically, we sample an indicator variable $z_c$ 
with the probability $\pi_c$ and, subsequently, we sample $\vec{s}$ from $\mathcal{N}(\vec{\mu}_{\vec{s},c},\vec{\Sigma}_{\vec{s}\vec{s},c})$. By using this sampling strategy, the input sequence $\{\vec{s}_n\}_{n=1}^{N}$ can be determined. In this way, the input sequence captures the probabilistic properties of the input space of demonstrations, where data-dense regions (hence important) will lead to more sampled input points and vice-versa for more sparse regions of the input space.
Accordingly, by using GMR we have the probabilistic reference trajectory with high-dimensional inputs, denoted by $\vec{D}_r^s=\{\vec{s}_n,\hat{\vec{\mu}}_n,\hat{\vec{\Sigma}}_n\}_{n=1}^{N}$. The resulting probabilistic reference trajectory $\vec{D}_r^s$ hence encapsulates the probabilistic features of demonstrations. It should be noted that, for sampling, $N$ needs not be the same as the number of points in each demonstration, hence one has the freedom to sub-sample when higher computational efficiency is required.

\subsection{Learning quaternions with high-dimension inputs\label{subsec:quat:learn:higDim}} 
	
Similarly to (\ref{equ:para:traj}), we formulate the parametric trajectory associated with high-dimensional inputs as
\begin{equation}
\vec{\xi}(\vec{s})
=\underbrace{\left[\begin{matrix} 
\vec{\phi}^{\trsp}(\vec{s}) & \vec{0} & \cdots &\vec{0} \\
\vec{0} & \vec{\phi}^{\trsp}(\vec{s}) &  \cdots &\vec{0} \\
\vdots & \vdots &  \ddots & \vdots \\
\vec{0} & \vec{0} &  \cdots & \vec{\phi}^{\trsp}(\vec{s}) \\
\end{matrix}\right]}_{\vec{\Theta}^{\trsp}(\vec{s})}\vec{w}.
\label{equ:para:traj:high}
\end{equation}
Note that $\frac{d \vec{\phi}(\vec{s})}{d t}$ is not included in (\ref{equ:para:traj:high}) since it relies on $\frac{d \vec{s}}{d t}$, which is often unpredictable in real applications. Formally, we formulate the imitation learning problem with multiple inputs as maximizing
\begin{equation}
J(\vec{w})=\prod_{n=1}^{N}\mathcal{P}(\vec{\Theta}^{\trsp}(\vec{s}_n)\vec{w}|\hat{\vec{\mu}}_n,\hat{\vec{\Sigma}}_n).
\label{equ:posterior:high}
\end{equation}
Consequently, we can follow (\ref{equ:post})--(\ref{equ:kernel:pred}) to derive the kernelized approach that is capable of learning high-dimensional inputs, except that the definition of $\vec{\phi}(\cdot)$ is different and thus the kernel definition in (\ref{equ:kernel}) becomes 
\begin{equation}
\vec{k}(\vec{s}_i,\vec{s}_j)
=\vec{\Theta}^{\trsp}(\vec{s}_i)\vec{\Theta}(\vec{s}_j)
=k(\vec{s}_i,\vec{s}_j) \vec{I}_6.
\label{equ:kernel:high}
\end{equation}
Therefore, given a query input $\vec{s}^{*}$,
we can employ (\ref{equ:kernel:pred}) to predict the output $\vec{\xi}(\vec{s}^{*})=\left[\begin{matrix} \vec{\xi}_{\vec{p}} \\ \vec{\xi}_{\vec{q}} \end{matrix}\right]$ and, subsequently, retrieve the corresponding Cartesian state through
$\left[\begin{matrix}
\vec{p}(\vec{s}^{*})\\ \vec{q}(\vec{s}^{*})
\end{matrix}\right]=
\left[\begin{matrix}
\vec{\xi}_{\vec{p}} \\ \exp(\vec{\xi}_{\vec{q}})* \vec{q}_a
\end{matrix}\right]$.

 \begin{figure*}[bt] \centering 
	\includegraphics[width=0.99\textwidth]{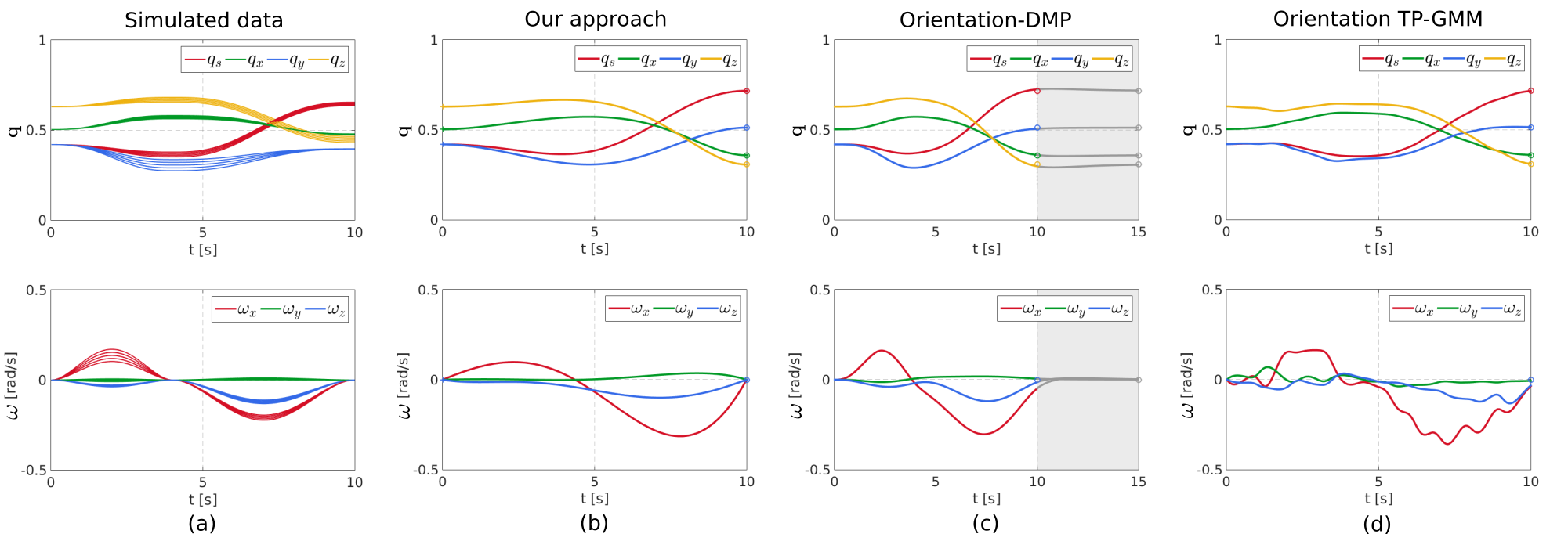}
	\caption{Evaluations of various approaches on simulated examples. (\emph{a}) shows simulated quaternion trajectories and their corresponding angular velocities. (\emph{b})--(\emph{d}) display adapted quaternion trajectories towards new target (i.e., end-point) as well as the adapted angular velocities by using  our approach (\emph{b}), orientation-DMP \cite{Ude,Abu} (\emph{c}) and orientation TP-GMM \cite{Joao} (\emph{d}). Note that for all approaches the desired movement duration is 10$s$, the shaded area in (\emph{c}) denotes extra time required for DMP. The circles with bright colors denote desired quaternions and angular velocities, while the gray circles in (\emph{c}) correspond to the delayed desired points. } 
	\label{fig:simu:eva} 
\end{figure*}

\begin{figure*}[bt] \centering 
	\includegraphics[width=0.99\textwidth]{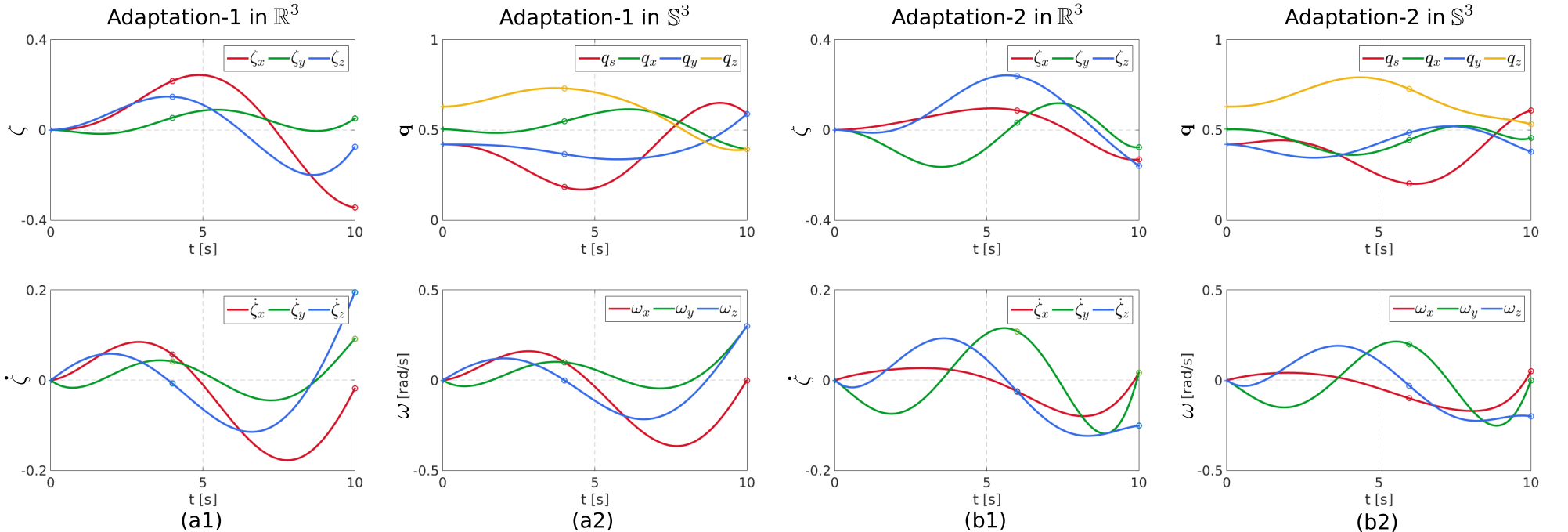}
	\caption{Adaptations of quaternion and angular-velocity profiles with various constraints of desired points (depicted by circles), where (\emph{a1})--(\emph{a2}) and (\emph{b1})--(\emph{b2}) correspond to the first and second evaluations, respectively. Note that (\emph{a1}) and (\emph{b1}) represent adaptations in the transformed space $\mathbb{R}^3$ that is determined via (\ref{equ:transfer}), while (\emph{a2}) and (\emph{b2}) correspond to adaptations in $\mathbb{S}^3$ space.} 
	\label{fig:simu:eva:adas} 
\end{figure*}

\subsection{Adapting quaternions with high-dimensional inputs\label{subsec:quat:ada:higDim}} 
Let us write $H$ desired points as $\tilde{\vec{D}}^{0}=\{\tilde{\vec{s}}_h,\tilde{\vec{\xi}}_h^{0}\}_{h=1}^{H}$ with
$\tilde{\vec{\xi}}_{h}^{0}=\left[\begin{matrix}
\tilde{\vec{p}}_{h} \\ \tilde{\vec{q}}_{h}
\end{matrix}\right]$.
Then, we transform the desired points $\tilde{\vec{D}}^{0}$ into Euclidean space, leading to $\tilde{\vec{D}}^s=\{\tilde{\vec{s}}_h,\tilde{\vec{\xi}}_h\}_{h=1}^{H}$ with 
$\tilde{\vec{\xi}}_{h}=\left[\begin{matrix}
\tilde{\vec{p}}_{h} \\ \log( \tilde{\vec{q}}_{h} * \bar{\vec{q}}_a)
\end{matrix}\right]$. In order to incorporate the adaptation precision, we can assign covariance matrices $\tilde{\vec{\Sigma}}_h$ for various transformed points $\tilde{\vec{\xi}}_h$. Thus, we have an additional reference trajectory that represents the transformed desired points in Euclidean space, i.e., $\tilde{\vec{D}}_r^s=\{\tilde{\vec{s}}_h,\tilde{\vec{\xi}}_h,\tilde{\vec{\Sigma}}_h\}_{h=1}^{H}$. According to the discussion in Section~\ref{subsec:ada:kmp}, we can concatenate $\tilde{\vec{D}}_r^s$ with $\vec{D}_r^s=\{\vec{s}_n,\hat{\vec{\mu}}_n,\hat{\vec{\Sigma}}_n\}_{n=1}^{N}$, resulting in an extended reference trajectory ${\vec{D}_r^{s}}^{U}$ that can be used to generate a 6-D trajectory\footnote{This trajectory passes through the transformed desired points $\tilde{\vec{D}}^s$.} in Euclidean space and later recover the Cartesian trajectory (comprising Cartesian position and quaternion) that passes through various desired points defined by $\tilde{\vec{D}}^{0}$.

It is worth mentioning that, given the joint probability distribution $\mathcal{P}(\vec{s},\vec{\xi})$ in (\ref{equ:gmm:highD}), GMR can be employed to predict the corresponding output for a query input $\vec{s}^{*}$ through calculating $\mathcal{P}(\vec{\xi}|\vec{s}^{*})$ . However, GMR is only limited for task reproduction (i.e., reproducing demonstrations). When the adaptation problem is encountered, e.g., the predicted trajectory must pass through the desired points  $\tilde{\vec{D}}^{0}$, GMR becomes inapplicable, since $\mathcal{P}(\vec{s},\vec{\xi})$ (extracted from demonstrations) does not address the constraints from desired points. 
In contrast, within our framework, both reproduction and adaptation issues can be directly tackled by 
using $\vec{D}_r^s$ or ${\vec{D}_r^{s}}^{U}$.

\section{Evaluations \label{sec:eva}}

In this section, we report several examples to illustrate the performance of our approach:  
\begin{enumerate}
	\item[](\emph{i}) orientation adaptation towards a desired target point (Section~\ref{subsec:comp}), where orientation-DMP \cite{Ude, Abu} and orientation TP-GMM \cite{Joao} are employed as comparisons\footnote{In this section, $\vec{q}$ is represented as $\vec{q}=[q_s \; q_x \; q_y \;q_z]^{\trsp}$.};
\item[](\emph{ii}) orientation adaptations towards arbitrary desired points in terms of quaternions and angular velocities (Section~\ref{subsec:eva:ada:simu});  
\item[](\emph{iii}) orientation adaptations with angular acceleration constraints (Section~\ref{subsec:jerk:ada:simu});
\item[](\emph{iv}) rhythmic orientation reproduction and adaptations (Section~\ref{subsec:eva:rhythmic:ada:simu});
\item[](\emph{v}) concurrent adaptations of Cartesian position and orientation in a painting task (Section~\ref{sec:real:eva});
\item[](\emph{vi}) learning Cartesian trajectory with high-dimensional inputs in a human-robot collaboration scenario (Section~\ref{sec:real:eva:highD}). 
\end{enumerate} 
The evaluations (\emph{i})--(\emph{iv}) are verified in 
simulated examples while (\emph{v})--(\emph{vi}) are carried out on real robots. Videos of the experimental evaluations as well as didactic codes are provided at \emph{https://sites.google.com/view/quat-kmp}.

\subsection{Evaluations on quaternion adaptations \label{subsec:simu}}

We collected five simulated quaternion trajectories with time-length $10s$, as depicted in Fig.~\ref{fig:simu:eva}(\emph{a}), where minimal jerk polynomial and renormalization are used to generate smooth and proper quaternion trajectories. 
In order to show the performance of our approach, we first compare it with orientation-DMP \cite{Ude, Abu} and orientation TP-GMM \cite{Joao} in Section~\ref{subsec:comp}. Subsequently, we evaluate our approach by adapting quaternions and angular velocities towards various desired points in Section~\ref{subsec:eva:ada:simu}. 
The Gaussian kernel $k(t_i,t_j)=\exp(-\ell (t_i-t_j)^2)$ with $\ell=0.01$ and the regularization factor $\lambda=1$ are used in this section.

\begin{table}[bt]
	\vspace{0.6cm}
	\caption {Planned Errors of Our Approach and State-of-the-Art}
	\centering
	\scalebox{0.85}
	{\begin{tabular}{lcccc}
			\toprule %
			& \multicolumn{2}{c}{Quaternion distance error$^{*}$} & \multicolumn{2}{c}{Angular velocity error} 
			\\
			&$t=10s^{**}$ & $t=15s$&$t=10s$ & $t=15s$ \hspace{0cm} 
			\\ 
			\toprule %
			Our approach 
			& 0
			& -  
			& 0.0017 
			& -
			\\
			\midrule
			Orientation--DMP \cite{Ude, Abu}
			& 0.0285
			& 0.0046
			& 0.0513
			& 0.0034
			\\
			\midrule
			Orientation TP-GMM \cite{Joao}
			& 0.0085
			& -  
			& 0.0498 
			& -	
			\\
			\bottomrule						
	\end{tabular}}
	\begin{tablenotes}
		\item[] {*\emph{Quaternion distance is calculated by} \cite{Ude,Abu}: 
			\begin{equation*}
			d(\vec{q}_1,\vec{q}_2)=
			\left\{ 
			\begin{aligned}
			&2\pi,\quad\quad\quad\quad\quad\quad\;\;\vec{q}_1*\bar{\vec{q}}_2=[-1\,0\,0\,0]^{\trsp}\\
			&2||\log(\vec{q}_1*\bar{\vec{q}}_2)||,\; otherwise.
			\end{aligned}  
			\right.
			\end{equation*}}
		\item[] {\emph{**Note that the desired movement duration is $10s$.}}
	\end{tablenotes}
	\label{table:error}
\end{table}

\begin{table}[bt]
	\vspace{0.6cm}
	\caption {Smoothness Costs of Our Approach and State-of-the-Art}
	\centering
	\scalebox{0.85}
	{\begin{tabular}{lcccc}
			\toprule %
			& Quaternion smoothness  & Angular-velocity smoothness 	
			\\
			&cost $c_q$ & cost $c_\omega$
			\\ 
			\toprule %
			Our approach
			& 7.4326 $\times$ $10^{-4}$
			& 8.5820 $\times$ $10^{-4}$
			\\
			\midrule
			Orientation--DMP \cite{Ude, Abu}
			& 7.7441 $\times$ $10^{-4}$
			& 9.3905 $\times$ $10^{-4}$
			
			\\
			\midrule
			Orientation TP-GMM \cite{Joao}
			& 7.8706 $\times$ $10^{-4}$
			& 1.6463 $\times$ $10^{-3}$
			\\
			\bottomrule						
	\end{tabular}}
	\begin{tablenotes}
		\item[] \emph{*The smoothness costs (\ref{equ:smooth:q})$-$(\ref{equ:smooth:w}) are evaluated between $0 s$ and $10s$.}
	\end{tablenotes}
	\label{table:smooth}
\end{table}

\subsubsection{Comparison with state-of-the-art approaches\label{subsec:comp}}
Since orientation-DMP is restricted to target (i.e., end-point) adaptation while having zero angular velocity at the ending point, we here 
consider an example with the desired point being
$\tilde{t}_1=10s$, $\tilde{\vec{q}}_1=[0.7172\;    0.3586\;    0.5123\;    0.3074]$, $\tilde{\vec{\omega}}_1=[0 \;0\;0]$. The auxiliary quaternion $\vec{q}_{a}$ is set as the initial value of simulated quaternion trajectories. 

The evaluations of using our approach and orientation--DMP are provided in Fig.~\ref{fig:simu:eva}(\emph{b})--(\emph{c}). It can be seen from Fig.~\ref{fig:simu:eva}(\emph{b}) that our approach is capable of generalizing learned quaternion trajectories to the new target point $\tilde{\vec{q}}_1$ while having zero angular velocity at the ending time $\tilde{t}_1$. However, orientation--DMP needs extra time (depicted by shaded area) to converge to the desired point. 
Furthermore, we use orientation TP-GMM\footnote{A second-order linear dynamics model is employed together with TP-GMM towards obtaining smooth trajectories. Please refer to \cite{Joao} for implementation details.} to tackle the same target adaptation problem, whose adapted trajectories are shown in Fig.~\ref{fig:simu:eva}(\emph{d}).

\begin{table}[bt]
	\vspace{0.6cm}
	\caption {Angular Acceleration Costs Under Different $\lambda_a$}
	\centering
	\scalebox{1.0}
	{\begin{tabular}{cccccc}
			\toprule %
			$\lambda_a$ 
			& $10^1$
			& $10^2$
			& $10^3$
			& $10^4$
			& $10^5$
			\\
			\midrule
			$c_{\omega d}$
			& 0.0307    
			& 0.0303      
			& 0.0273        
			& 0.0183     
			& 0.0140    
			\\
			\bottomrule						
	\end{tabular}}
	\label{table:cost}
\end{table}

\begin{figure*}[bt] \centering 
	\includegraphics[width=0.95\textwidth]{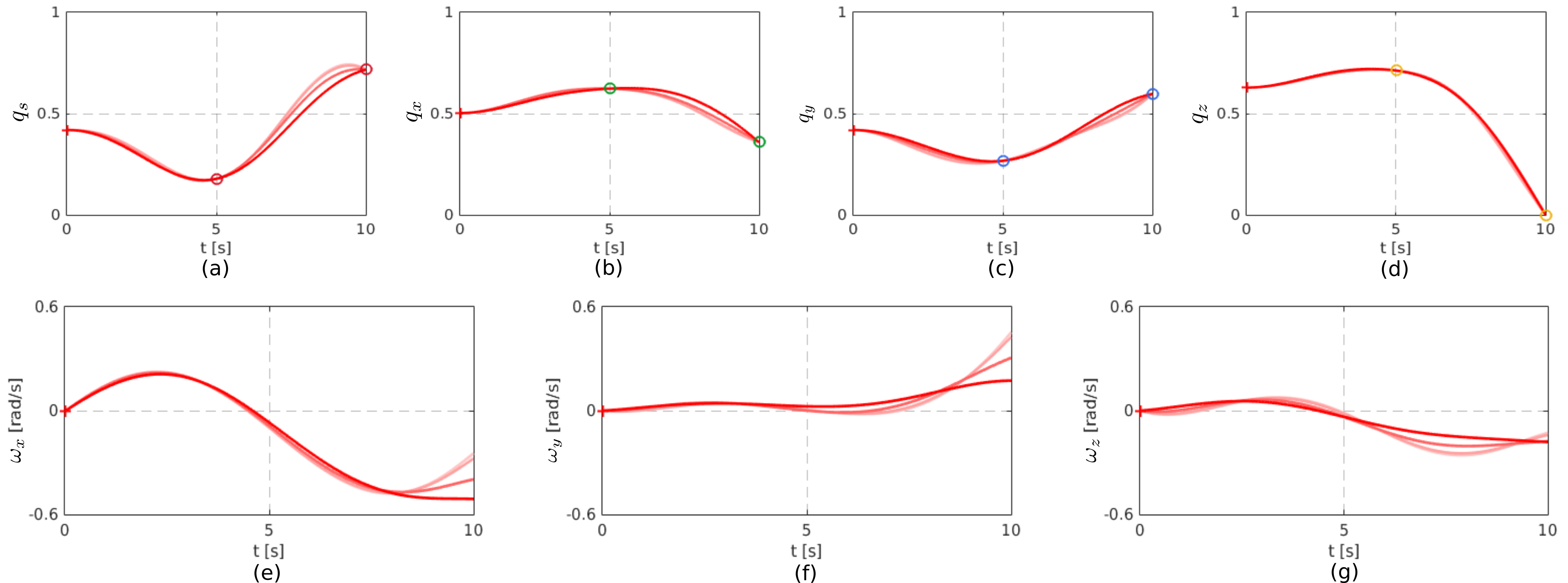}
	\caption{Orientation adaptations with angular acceleration constraints, where quaternion profiles are adapted towards various desired points (depicted by colorful circles). The cross `+' represents the starting point of trajectories. The solid curves correspond to different values of $\lambda_a$, with color that switches from light red to dark red corresponding to the increasing direction of $\lambda_a$.} 
	\label{fig:simu:eva:jerk:evolve} 
\end{figure*} 

\begin{figure*}[bt] \centering 
	\includegraphics[width=0.95\textwidth]{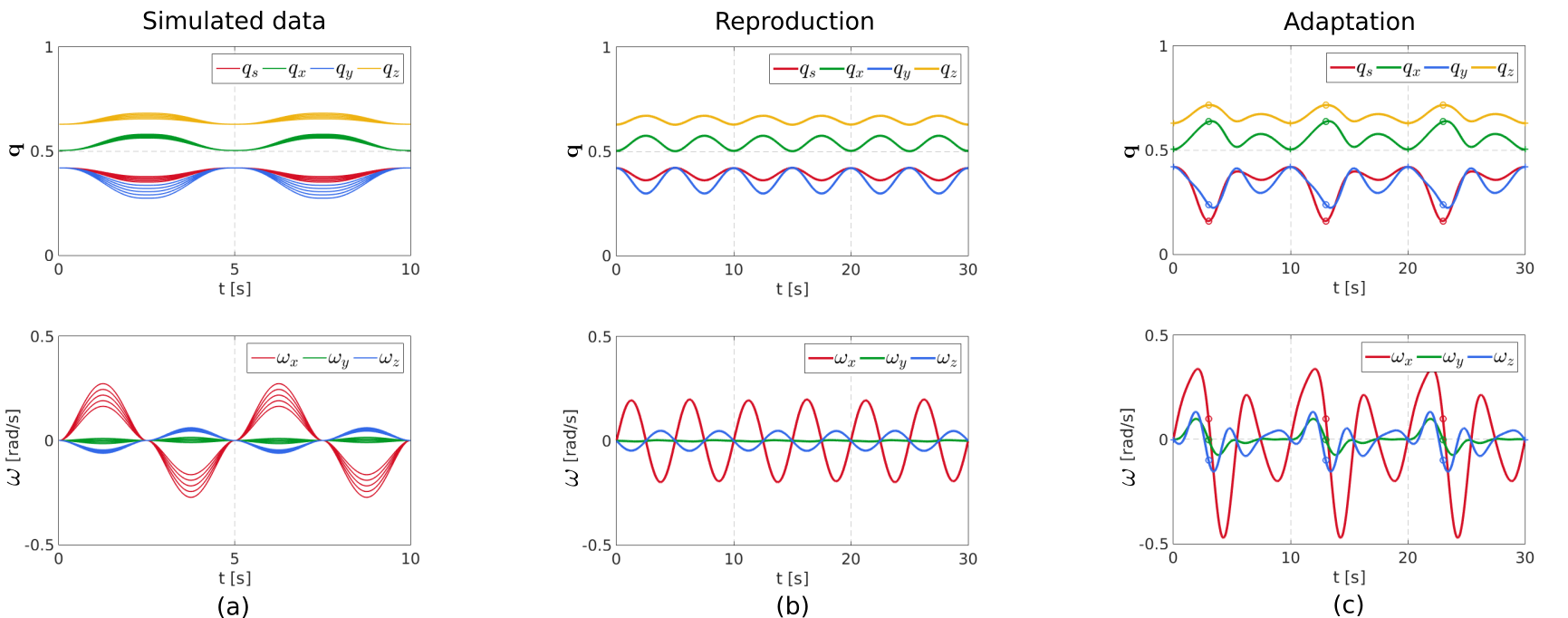}
	\caption{Reproduction and adaptation of rhythmic orientation trajectories by using our approach. (\emph{a}) plots simulated quaternions and their corresponding angular velocities, where the motion period is $10s$. (\emph{b}) and (\emph{c}) show orientation reproduction and adaptation over three periods (i.e., $30s$), respectively. Circles in (\emph{c}) denote the desired quaternion and angular velocity.}
	\label{fig:simu:eva:thythmic} 
\end{figure*}

The planned errors of three methods in comparison with the desired point is summarized in Table~\ref{table:error}, showing that our approach achieves the best performance in terms of adaptation precision. Note that the errors from DMP can possibly be further reduced by tuning the relevant parameters, e.g., the number of basis functions, bandwidth of basis functions and the length of time step, but here we present the best results we could obtain.
In addition, we evaluate the smoothness of adapted quaternion and angular-velocity profiles, where the smoothness cost for quaternion is defined as
\begin{equation} c_{q}=\frac{1}{N}\sum_{n=1}^{N-1}||\vec{q}(t_{n+1})-\vec{q}(t_n)||
\label{equ:smooth:q}
\end{equation}
	 and the cost for angular velocity is
	 \begin{equation}
	  c_{\omega}=\frac{1}{N}\sum_{n=1}^{N-1}||\vec{\omega}(t_{n+1})-\vec{\omega}(t_n)||.
\label{equ:smooth:w}
	  \end{equation}
As can be seen in Table~\ref{table:smooth}, our approach corresponds to the smallest costs in terms of both $c_{q}$ and $c_{\omega}$.

\subsubsection{Adapting quaternion trajectory towards various desired points\label{subsec:eva:ada:simu}}
Now, we consider a more challenging adaptation task that needs various desired points (i.e., via-/end- points) in terms of quaternion and angular velocity. Note that orientation-DMP \cite{Ude, Abu} and orientation TP-GMM \cite{Joao} are not applicable in this case. Two groups of quaternion adaptations in $\mathbb{S}^3$, accompanied by the corresponding adaptations of the projected trajectories via (\ref{equ:transfer}) in $\mathbb{R}^3$, are shown in Fig.~\ref{fig:simu:eva:adas}, showing that our approach indeed modulates quaternions and angular velocities to pass through various desired points (plotted by circles).

\subsection{Evaluations on quaternion adaptations with angular acceleration constraints\label{subsec:jerk:ada:simu}}
In this section, we consider quaternion adaptations with angular acceleration constraints, where the same simulated demonstrations, as plotted in Fig.~\ref{fig:simu:eva}(\emph{a}), are employed. Specifically, we aim to adapt quaternion profile,
while taking into account the angular acceleration constraints. In order to quantitatively show the performance of our approach, we define the angular acceleration cost as 
\begin{equation}
c_{\omega d}=\frac{1}{N}\sum_{n=1}^{N}||\dot{\vec{\omega}}(t_n)||^2
\end{equation}
and meanwhile a group of penalty parameters $\lambda_a$ are used. We use Gaussian kernel for the evaluations. Other relevant parameters are set as $\ell=0.01$ and $\lambda=1$. 

The evolved trajectories of quaternion and angular velocity are depicted in Fig.~\ref{fig:simu:eva:jerk:evolve}, where the color changes from light to dark as $\lambda_a$ increases. Note that the evolved trajectories pass through various desired points (depicted by circles) precisely. The corresponding angular acceleration costs with different $\lambda_a$ are provided in Table~\ref{table:cost}, representing that $c_{\omega d}$ decreases as $\lambda_a$ increases (which indeed coincides with our interpretation of the penalty coefficient). Thus, 
we can conclude that our approach is capable of adapting quaternions towards various desired points while incorporating angular acceleration constraints.

\subsection{Evaluations on rhythmic quaternion trajectories\label{subsec:eva:rhythmic:ada:simu}}

Differing from the aforementioned examples on point-to-point quaternions, we here test our approach on rhythmic quaternion trajectories. Note that rhythmic quaternions are very important in many orientation-sensitive tasks, such as screwing a lid off the bottle and wiping a surface. Similarly to Section~\ref{subsec:simu}, we use polynomials to generate five demonstrations (each lasts for $10s$) for training our approach, as shown in Fig.~\ref{fig:simu:eva:thythmic}(\emph{a}). The periodic kernel \cite{Duven} $k(t_i,t_j)=\exp(-\ell sin^2(\frac{t_i-t_j}{T}\pi))$  with $\ell=0.4$ and $T=10s$ is employed.
The regularization factor is set to be $\lambda=10$. In this section, the angular acceleration constraints are not considered, i.e., $\lambda_a=0$, but one can easily incorporate these constraints into rhythmic movements.

We first consider the reproduction capability of our approach, where quaternion and angular-velocity profiles over three periods (i.e., $30s$) are generated. It can be seen from Fig.~\ref{fig:simu:eva:thythmic}(\emph{b}) that our method can reproduce trajectories that maintain the shape of demonstrations and meanwhile exhibit rhythmic properties. Second, we test the adaptation capability of our method by imposing a via-point constraint at $t=3s$. The adapted trajectories over three periods are given in Fig.~\ref{fig:simu:eva:thythmic}(\emph{c}), where the quaternion and angular-velocity trajectories are modulated towards the via-point (depicted by circles) in each period. Moreover, the rhythmic property is kept in this adaptation case.

 \begin{figure}[bt] \centering
	\includegraphics[width=0.48\textwidth]{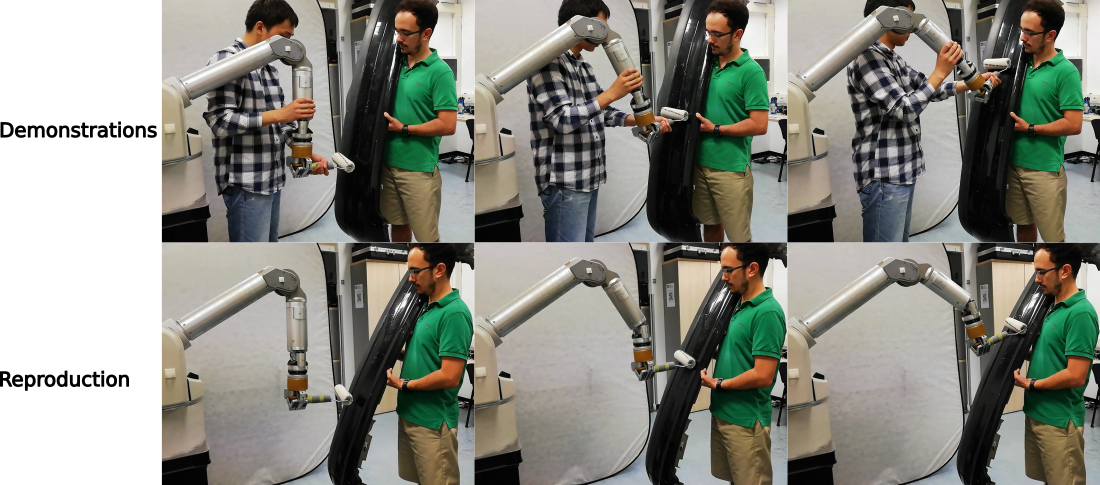}
	\caption{Painting task on the real Barrett WAM robot. \emph{First row} shows kinesthetic teaching of the painting task. \emph{Second row} represents the task reproduction. }
	\label{fig:real:robot} 
\end{figure}

\begin{figure*}[bt] \centering 
	\includegraphics[width=0.97\textwidth]{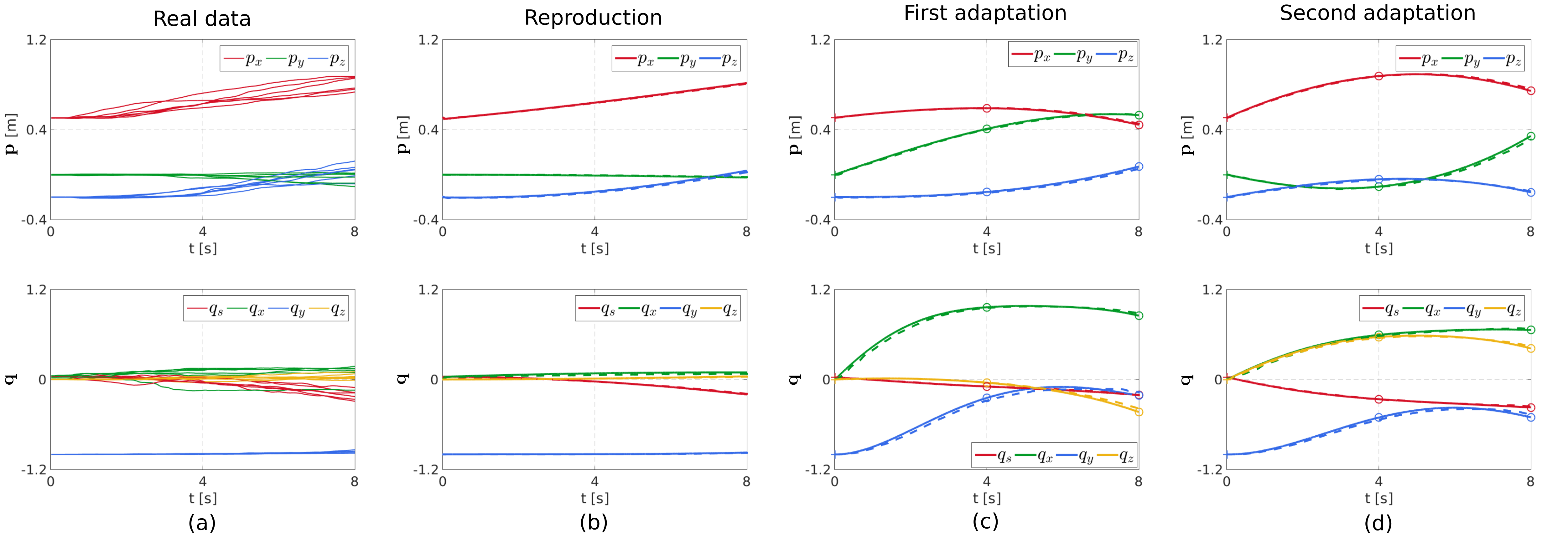}
	\caption{Evaluations of our approach through a painting task on the real Barrett WAM robot. (\emph{a}) shows demonstrated Cartesian positions and quaternions in the painting task. (\emph{b}) represents Cartesian trajectory in task reproduction. (\emph{c})--(\emph{d}) correspond to adapted Cartesian trajectories with various desired points. In plots (\emph{b})--(\emph{d}), solid curves represent planned trajectories by using our approach while dashed curves denote real measured trajectories. Circles depict desired Cartesian positions and quaternions. } 
	\label{fig:real:curves} 
\end{figure*}

\subsection{Evaluations of learning time-driven Cartesian trajectories on real robot \label{sec:real:eva}}

We here consider a painting task that requires the real Barrett WAM robot to paint different areas with proper orientations. 
Through kinesthetic teaching (\emph{first row} in Fig.~\ref{fig:real:robot}), six demonstrations comprising time, Cartesian position and quaternion are recorded, as shown in Fig.~\ref{fig:real:curves}(\emph{a}). We first apply our approach to task reproduction, i.e., without incorporating the desired points. The auxiliary quaternion $\vec{q}_{a}$ is set as the initial value of demonstrations. Gaussian kernel with $\ell=0.001$ is used and $\lambda=1$. 
The reproduced  Cartesian trajectory
(solid curves) and its corresponding measured trajectory (dashed curves) are shown in Fig.~\ref{fig:real:curves}(\emph{b}). Snapshots of task reproduction are provided in Fig.~\ref{fig:real:robot} (\emph{second row}), where we can see that the robot is capable of reproducing a similar task to that demonstrated by the human.

Now, we consider two groups of adaptation evaluations and, in each group, demonstrated Cartesian trajectories are modulated towards two unseen desired points (i.e., via-point and end-point). 
The adapted Cartesian trajectories are shown in Fig.~\ref{fig:real:curves}(\emph{c})-(\emph{d}), where the planned trajectories (solid curves) and real measured trajectories (dashed curves) are provided. 
It can be seen that the planned trajectories are capable of meeting various constraints, i.e., Cartesian position and quaternion constraints.
More explanations of the adaptation evaluations are provided in our previous work \cite{Huang2019_1}.

\begin{figure}[t] \centering
	\includegraphics[width=0.48\textwidth]{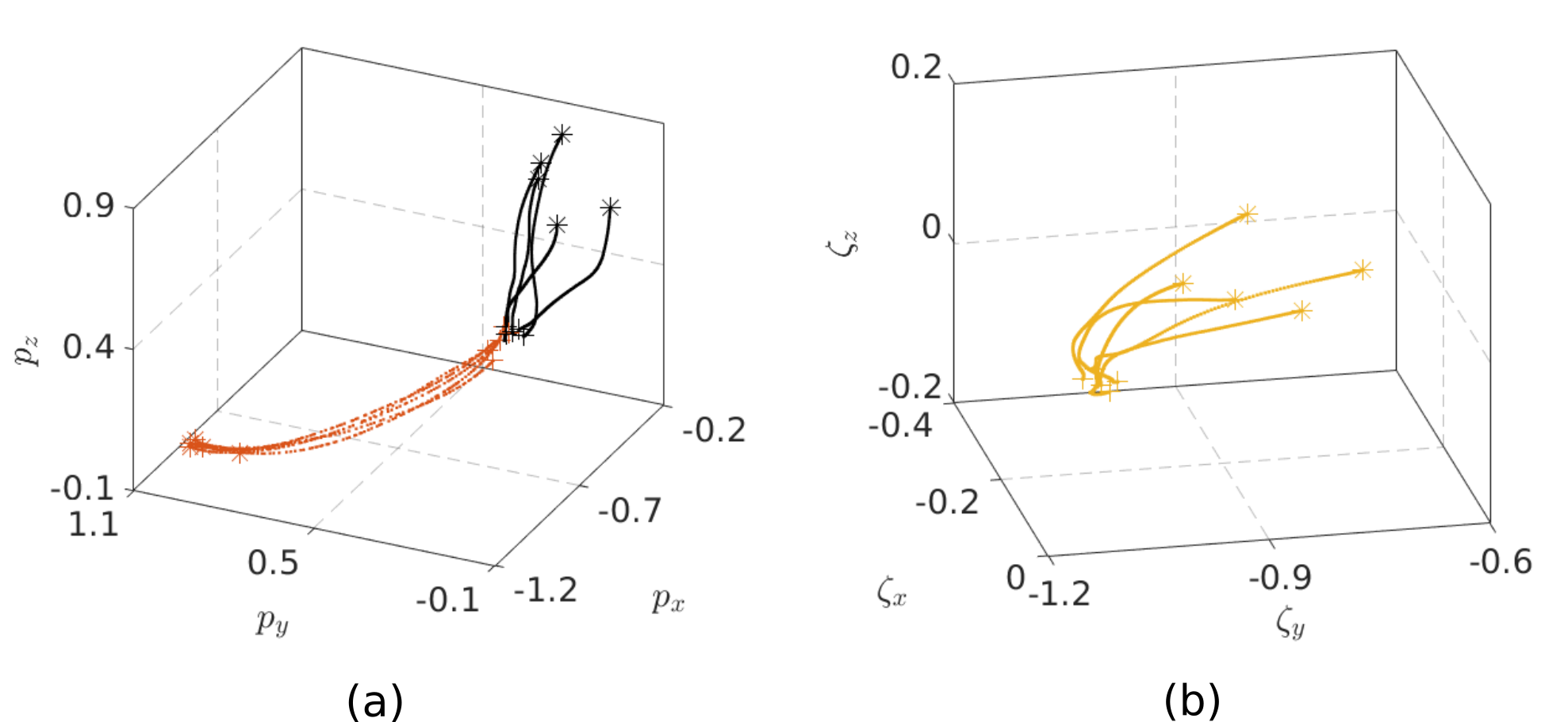}
	\caption{Demonstrated human hand trajectories and robot Cartesian trajectories in a human-robot handover task. (\emph{a}) shows human hand positions (red curves) and robot Cartesian positions (black curves). For the sake of visualization, the transformed trajectories of robot quaternions are plotted in (\emph{b}).}
	\label{fig:handover:demos} 
\end{figure}

\begin{figure}[t] \centering
	\includegraphics[width=0.48\textwidth]{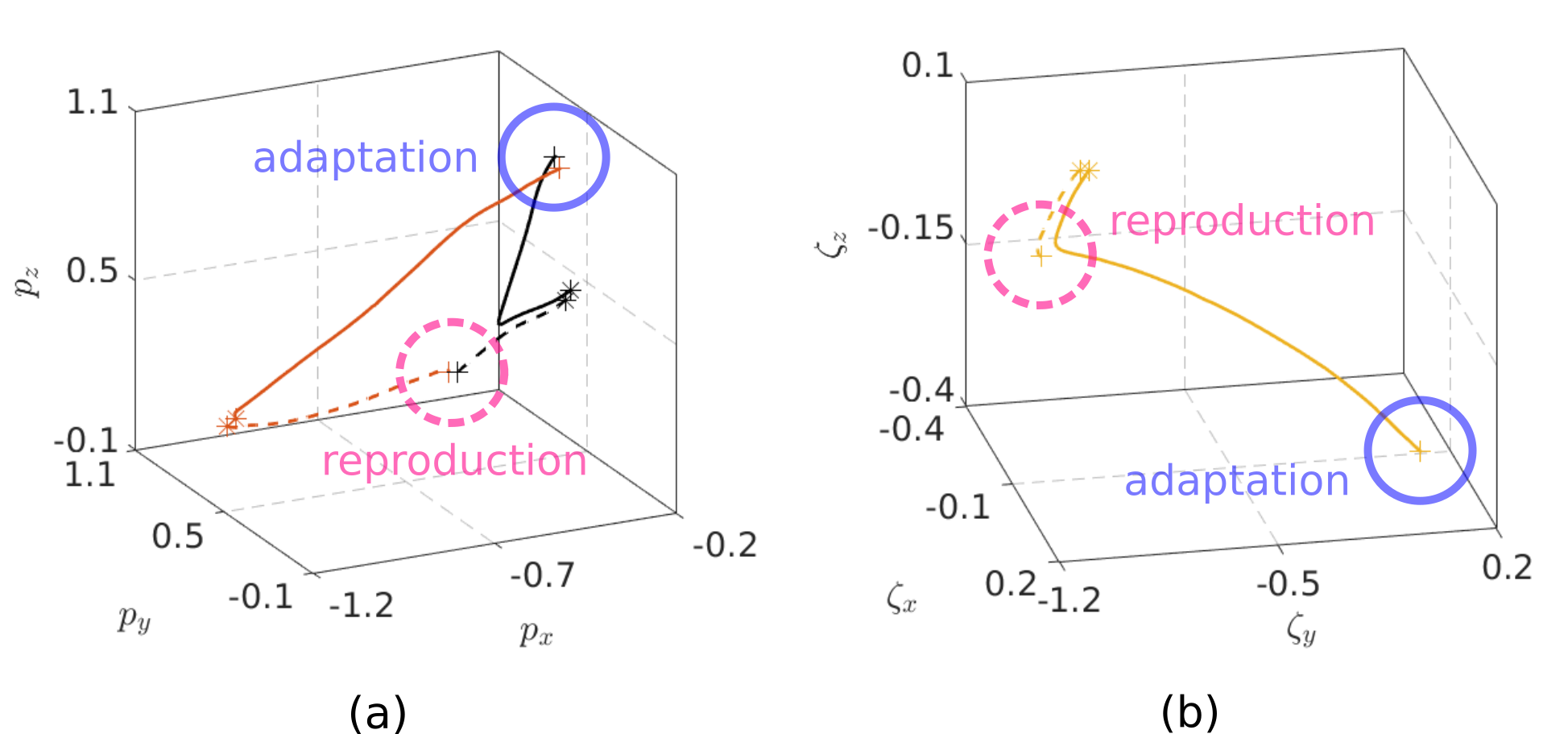}
	\caption{Reproduction (dashed curves) and adaptation (solid curves) evaluations in the handover task. (\emph{a}) represents human hand positions (red curves) and robot Cartesian positions (black curves). (\emph{b}) plots the transformed trajectories of robot quaternions via (\ref{equ:transfer}).} 
	\label{fig:handover:eva} 
\end{figure}

\begin{figure}[t] \centering
	\includegraphics[width=0.485\textwidth]{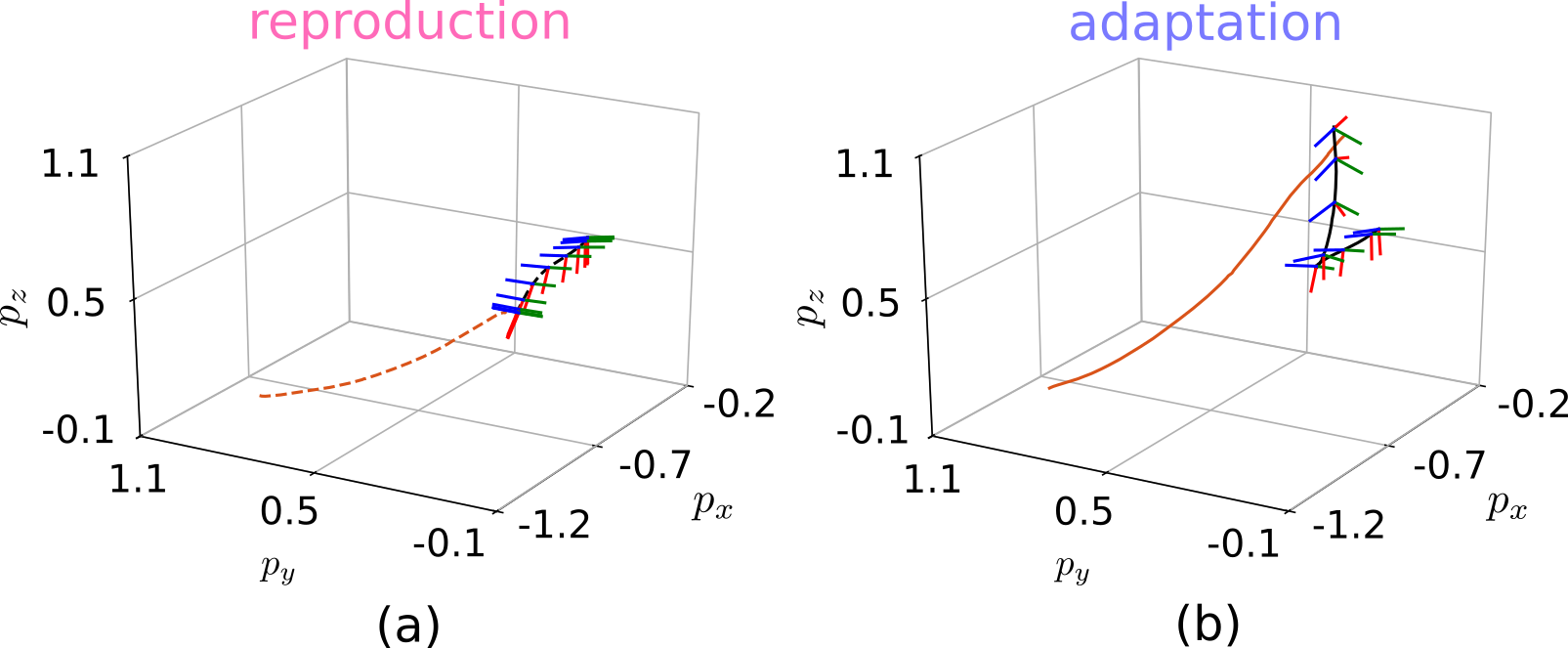}
\caption{	Illustration of human hand trajectory and robot Cartesian trajectory in reproduction (\emph{a}) and adaptation (\emph{b}) evaluations, where the red curves denote human hand positions, the black curves denote the end-effector positions and frames correspond to its orientations.}
	\label{fig:handover:eva:together} 
\end{figure}

\subsection{Evaluations of learning Cartesian trajectories with high-dimensional inputs on real robot \label{sec:real:eva:highD}}

In this section, we consider a human-robot handover task, where the robot moves towards the human user in order to accomplish the handover task. Specifically, we consider human hand position\footnote{An optical tracker is used to measure human hand position.} as inputs while robot end-effector position and orientation as outputs. It is worth emphasizing that we aim to predict robot Cartesian state (6-D\footnote{We here refer to quaternion as a 3-D variable due to the norm constraint, albeit that our approach predicts four elements of quaternion simultaneously.}) in accordance to human hand state (3-D) directly, without any additional operations like phase-estimation \cite{Amor,Maeda}. 

We collect five demonstrations in terms of human hand position (red curves), as well as robot end-effector position (black curves) and quaternion, as show in Fig.~\ref{fig:handover:demos}. Note that the transformed trajectories of quaternions (yellow curves) via (\ref{equ:transfer}) are plotted for the sake of visualization. Then, by following the description in Section~\ref{subsec:quat:model:higDim}, we can generate a reference trajectory (associated with high-dimensional inputs) for training our approach. 
The auxiliary quaternion $\vec{q}_{a}$ is set to be $\vec{q}_{a}=[1 \, 0\, 0 \, 0]^{\trsp}$. 
The Gaussian kernel is used and the related parameters are $\ell=1$ and $\lambda=2$.

In order to evaluate our approach, we first consider a reproduction task and subsequently an adaptation task where a new handover location is needed.
Figure~\ref{fig:handover:eva} depicts human hand trajectory (dashed red curve) and the corresponding robot Cartesian trajectory planned by our approach (dashed black and yellow curves) in the reproduction case, where 
Fig.~\ref{fig:handover:eva}(\emph{a}) and Fig.~\ref{fig:handover:eva}(\emph{b}) correspond to positions and transformed quaternion data, respectively.  
In addition, in Fig.~\ref{fig:handover:eva:together}(\emph{a}) the human hand positions, robot Cartesian positions and orientations (in terms of frames) are depicted together.
It can be seen that the robot accomplishes the handover task
when human hand trajectory resembles the demonstrated ones. 

\begin{figure}[bt] \centering
	\includegraphics[width=0.49\textwidth]{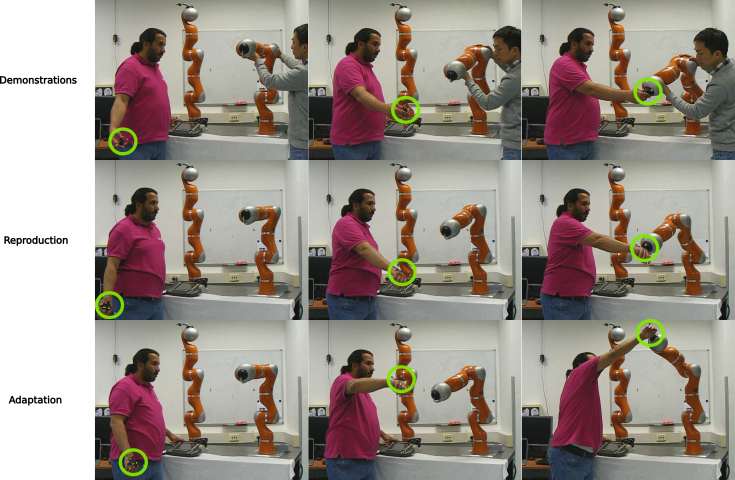}
	\caption{Handover task on the real KUKA robot. \emph{First row} shows kinesthetic teaching of the handover task. \emph{Second row} and \emph{third row} represent the reproduction and adaptation evaluations, respectively. Green circle corresponds to the optical tracker which measures the user's hand position on-line.} 
	\label{fig:handover:snapshot} 
\end{figure}

Now, we apply our approach to the adaptation situation, where the handover takes place at a new point that is unseen in demonstrations. This adaptation can be achieved by adding a desired point $\{\tilde{\vec{s}}_1,\tilde{\vec{p}}_{1},\tilde{\vec{q}}_{1}\}$ into the original reference trajectory, where $\tilde{\vec{s}}_1=\tilde{\vec{p}}_{1}=\vec{p}^{new}$ and $\tilde{\vec{q}}_{1}=\vec{q}^{new}$, ensuring that the robot reaches the new handover location $\vec{p}^{new}$ with desired quaternion $\vec{q}^{new}$ when the human hand arrives at $\vec{p}^{new}$. The adaptation evaluation is provided in Fig.~\ref{fig:handover:eva}, where the solid red curve denotes the user hand trajectory while the solid black and yellow curves correspond to the planned Cartesian trajectory for the robot. Again, we here only provide the transformed data of quaternions for the sake of easy observation.  
Similarly to the reproduction case, we represent human hand positions, robot Cartesian positions and orientations (in terms of frames) into a single plot, as shown in Fig.~\ref{fig:handover:eva:together}(\emph{b}).
By observing Fig.~\ref{fig:handover:eva} and \ref{fig:handover:eva:together}(\emph{b}), we can find that robot trajectory is indeed modulated according to the user hand position, leading to a successful handover at the new location. 
Snapshots of kinesthetic teaching of handover task as well as reproduction and adaptation evaluations are shown in Fig.~\ref{fig:handover:snapshot}. 
Thus, our approach is effective in both reproduction and adaptation cases while considering high-dimensional inputs.

\section{Discussion\label{sec:discuss}}

In this section, we discuss some related work on learning time-driven demonstrations with via-point constraints, as well as learning demonstrations comprising high-dimensional inputs  (Section~\ref{subsec:related:work}). Then, we discuss limitations and possible extensions of our approach (Section~\ref{subsec:limitation}). 

\subsection{Related work \label{subsec:related:work}}

The topic of via-point adaptation has been the focus of a few works in the imitation learning literature, e.g., \cite{Paraschos,Stulp,Weitschat}. In \cite{Paraschos}, Gaussian conditioning operation was used in ProMP to address the via-point issue. In \cite{Stulp}, the task-parameterized DMP was studied, where the via-point constraint was handled as a task parameter vector. However, both \cite{Paraschos} and \cite{Stulp} did not take into account the Cartesian orientation. Please note that \cite{Stulp} in essence focuses on learning time-driven demonstrations (i.e., demonstrations comprising 1-D time input), albeit that the task parameters that describe the condition of demonstrations could be high-dimensional.  Similarly, within the DMP framework, a time-varying target function was formulated in \cite{Weitschat} to incorporate via-points in terms of Cartesian position, while the orientation was missing.

In order to cope with the via-quaternion and via-angular velocity constraints, the strategy of sequencing different DMPs was proposed in \cite{Saveriano}. To take the sequence of two DMPs as an example, in \cite{Saveriano} the first DMP was used to plan a trajectory from the starting quaternion to the via-quaternion, and subsequently the second DMP was used to generate a trajectory from the via-quaternion to the target quaternion. Note that these two DMPs were trained by using different parts of a demonstration. Differing from \cite{Saveriano} where the demonstration needs to be segmented for each movement primitive, we propose to learn and adapt the entire demonstrations  
using a single movement primitive, where the segmentation of demonstrations and the sequence of multiple motion primitives are not required. Specifically, in contrast to \cite{Stulp,Weitschat,Saveriano} where each DMP corresponds to a single training trajectory, we study imitation learning from a probabilistic perspective and exploit the consistent features underlying multiple demonstrations. As a result, including via-points in our approach can be done in a rather straightforward way by simply defining the via-point, its associated input and the desired precision in the form of a covariance matrix.

Various imitation learning approaches (e.g., DMP \cite{Ijspeert} and ProMP \cite{Paraschos}) have been employed in human-robot collaboration \cite{Amor,Maeda}. However, the majority of these works model human motion and robot motion with time (i.e., learning demonstrations with time input), which will lead to the synchronization issue (see Section~\ref{subsec:quat:moti:higDim}). 
Note that DMP and ProMP explicitly depend on basis functions, which are non-trivial to extend to learn demonstrations comprising high-dimensional inputs, due to the 
\emph{curse of dimensionality}. As discussed in \cite{Bishop}, the number of basis functions often increases significantly as the dimension of inputs increases. In addition, the process of defining proper parameters for basis functions over high-dimensional state is cumbersome. For instance, the definition of a multivariate Gaussian basis function requires a center vector and a covariance matrix. In contrast, our approach introduces the kernel trick (see (\ref{equ:kernel:high})) and thus basis functions are not needed, yielding a non-parametric solution.

Moreover, note that GMM/GMR based approaches \cite{Kim} are capable of learning demonstrations comprising high-dimensional inputs, whereas the adaptation feature is not provided. In order to endow GMM/GMR with the adaptation capability, the task parameterized treatment (i.e., TP-GMM) was studied in \cite{Joao,Matin}. However, TP-GMM suffers from two main limitations: (\emph{i}) for the case of adapting demonstrations with multiple inputs, TP-GMM is restricted to target adaptation, where the via-point adaptation is not allowed; (\emph{ii}) for the case of adapting demonstrations with time input, TP-GMM is unable to deal with the via-point and angular-velocity constraints (see also the discussion in Section~\ref{subsec:eva:ada:simu}). 
Unlike TP-GMM, our approach can handle the adaptation issue with arbitrary desired points (including via-/end- points), as well as the angular velocity constraints.

Finally, it is worth mentioning that, as an extension of \cite{Ude,Abu}, in \cite{Kramberger} the orientation DMP was used for generalizing demonstrations towards different contextual (or conditional, task-specific) states.
Specifically, in the training phase, each fixed contextual state\footnote{The contextual state in \cite{Kramberger} can be interpreted as the condition under which the whole time-driven demonstration has taken place.} corresponds to one demonstration that consists of a time sequence (i.e., 1-D input) and its associated robot trajectory (i.e., outputs), and in the evaluation phase, the new contextual state is used to predict the corresponding DMP parameters that, subsequently, can be used to generate the entire robot trajectory. Thus, the work in \cite{Kramberger} can be viewed as learning of demonstrations with time input, while considering additional contextual states.  
Differing from \cite{Kramberger},
in our framework (see Section~\ref{sec:ori:highDim}), each demonstration comprises a high-dimensional varying input trajectory 
and a robot trajectory, and in the evaluation phase, the high-dimensional inputs are employed to directly predict the corresponding robot actions (see the hand-over experiment, where the 3-D user's hand position is utilized to predict the 6-D robot Cartesian state).
As we have shown, our approach alleviates the synchronization issue, allowing the robot to react in run-time to the human behavior that changes arbitrarily (e.g. with different speed). The adaptation capabilities of our approach complement this feature by allowing the robot to react, even to human behaviors that were not explicitly shown.

\subsection{Limitations and extensions of our approach \label{subsec:limitation}}

As explained in Section~\ref{sec:model:demo}, the learning and adaptation of quaternions are carried out in Euclidean space, where the $\log(\cdot)$ mapping in (\ref{equ:quat:diff}) is used to transform quaternions into Euclidean space. 
In order to guarantee that similar quaternions $\vec{q}$ correspond to similar $\vec{\zeta}$ in Euclidean space, we have imposed the Assumption 1. This assumption may restrict the applications of our method. For example, when the demonstrated quaternion trajectories differ from each other dramatically, the Assumption 1 could be violated, potentially invalidating the teaching of highly dynamic motions.

It is noted that we only focus on the prediction of quaternion profile 
in this paper. In fact, we can also predict the covariance (i.e., $3\times3$ matrix) of $\vec{\zeta}$ in $\mathbb{R}^3$, similarly to \cite{Huang2017}. Indeed, specifying a covariance matrix directly in $\mathbb{S}^3$ is not possible, with other state-of-the-art approaches following the direction of representing variability/correlation in Euclidean spaces with $3\times3$ matrix \cite{Kim,Matin}. Interestingly, some recent works on exploiting the covariance of trajectories and optimal control were reported, e.g., uncertainty-aware controller \cite{Joao2019} and minimum intervention controller \cite{Calinon2014}. Thus, it would be useful to integrate these ideas with our framework so as to perform orientation tasks in a safe and user-friendly way.
In addition, we set kernel parameters experimentally in our evaluations. Despite the definition of the parameters being relatively straightforward (intuition about the kernel width can be derived from the order magnitude of the inputs), as an extension of our work, we plan to provide a theoretical guidance for choosing kernel parameters.

\section{Conclusions\label{sec:conclusion}}
In this paper, we proposed an analytical approach for adapting quaternion and angular velocity towards arbitrary desired points.
In addition, our method is capable of incorporating angular acceleration or jerk constraints.  
In comparison with previous works (e.g., \cite{Ude, Abu}) that mostly focus on orientation adaptation towards target points, our work allows for broader applications, particularly when both quaternion and angular velocity need to be modulated.  Moreover, our approach is capable of learning quaternions associated with high-dimensional inputs (e.g., 3-D inputs were used in the real handover task), which is a quite desirable property in human-robot collaboration.

\section*{Acknowledgement}
We thank anonymous reviewers for their constructive and helpful comments on this paper.

\begin{appendices}

\section{Kernel Derivation Under Angular Acceleration Constraints \label{app:kernel}}
According to the definitions of $\vec{\Theta}(t)$ and $\vec{\varphi}(t)$, i.e., (\ref{equ:para:traj}) and (\ref{equ:const:notation}), we have

\begin{equation}
\begin{aligned}
&\vec{\Omega}^{\trsp}\!(t_i) \vec{\Omega}(t_j)=
\left[\!\begin{matrix}
\vec{\Theta}^{\trsp}(t_i){\vec{\Theta}}(t_j) & \vec{\Theta}^{\trsp}(t_i){\vec{\varphi}}(t_j)\\  
\vec{\varphi}^{\trsp}(t_i){\vec{\Theta}}(t_j) & \vec{\varphi}^{\trsp}(t_i){\vec{\varphi}}(t_j)
\end{matrix}\right]
\\
&=\left[\!\begin{matrix}
\vec{\phi}^{\trsp}(t_i){\vec{\phi}}(t_j)\vec{I}_{3} & \vec{\phi}^{\trsp}(t_i)\dot{\vec{\phi}}(t_j)\vec{I}_{3} & \vec{\phi}^{\trsp}(t_i)\ddot{\vec{\phi}}(t_j)\vec{I}_{3}\\ 
\dot{\vec{\phi}}^{\trsp}(t_i){\vec{\phi}}(t_j)\vec{I}_{3} & \dot{\vec{\phi}}^{\trsp}(t_i)\dot{\vec{\phi}}(t_j)\vec{I}_{3} & \dot{\vec{\phi}}^{\trsp}(t_i)\ddot{\vec{\phi}}(t_j)\vec{I}_{3}\\  
\ddot{\vec{\phi}}^{\trsp}(t_i){\vec{\phi}}(t_j)\vec{I}_{3} & \ddot{\vec{\phi}}^{\trsp}(t_i)\dot{\vec{\phi}}(t_j)\vec{I}_{3} & \ddot{\vec{\phi}}^{\trsp}(t_i)\ddot{\vec{\phi}}(t_j)\vec{I}_{3}
\end{matrix}\right] 
\label{equ:app:basis}
\end{aligned}
\end{equation}

It is well known that we can write $\vec{\phi}^{\trsp}(t_i){\vec{\phi}}(t_j)=k(t_i,t_j)$ \cite{Bishop}. However, when we calculate $\vec{\Omega}^{\trsp}(t_i)\vec{\Omega}(t_j)$ in (\ref{equ:app:basis}), the terms $\vec{\phi}^{\trsp}(t_i)\dot{\vec{\phi}}(t_j)$, $\vec{\phi}^{\trsp}(t_i)\ddot{\vec{\phi}}(t_j)$,
$\dot{\vec{\phi}}^{\trsp}(t_i){\vec{\phi}}(t_j)$, $\dot{\vec{\phi}}^{\trsp}(t_i)\dot{\vec{\phi}}(t_j)$,
$\dot{\vec{\phi}}^{\trsp}(t_i)\ddot{\vec{\phi}}(t_j)$,
$\ddot{\vec{\phi}}^{\trsp}(t_i){\vec{\phi}}(t_j)$,
$\ddot{\vec{\phi}}^{\trsp}(t_i)\dot{\vec{\phi}}(t_j)$,
$\ddot{\vec{\phi}}^{\trsp}(t_i)\ddot{\vec{\phi}}(t_j)$
are also encountered.
We here propose to approximate $\dot{\vec{\phi}}(t)$ and $\ddot{\vec{\phi}}(t)$, i.e.,
\begin{equation*}
\dot{\vec{\phi}}(t)
\approx \frac{{\vec{\phi}}(t+\delta)-{\vec{\phi}}(t)}{\delta} \, \mathrm{and} 
\end{equation*}
\begin{equation*}
\ddot{\vec{\phi}}(t)
\approx \frac{\dot{\vec{\phi}}(t+\delta)-\dot{\vec{\phi}}(t)}{\delta} \approx\frac{{\vec{\phi}}(t+2\delta)-2{\vec{\phi}}(t+\delta)+{\vec{\phi}}(t)}{\delta^{2}},
\end{equation*}
where $\delta>0$ denotes a small constant. 
By using these approximations, (\ref{equ:app:basis}) can be kernelized.

To take $\ddot{\vec{\phi}}^{\trsp}(t_i)\ddot{\vec{\phi}}(t_j)$ as an example, we have
\begin{equation*}
\begin{aligned}
&\ddot{\vec{\phi}}^{\trsp}(t_i)\ddot{\vec{\phi}}(t_j)=\biggl(\frac{{\vec{\phi}}^{\trsp}(t_i+2\delta)-2{\vec{\phi}}^{\trsp}(t_i+\delta)+{\vec{\phi}}^{\trsp}(t_i)}{\delta^{2}}\biggr)\\ &\quad\quad\quad\quad\quad\quad\quad\quad\quad \biggl(\frac{{\vec{\phi}}(t_j+2\delta)-2{\vec{\phi}}(t_j+\delta)+{\vec{\phi}}(t_j)}{\delta^{2}}\biggr)\\
&=\frac{\vec{\phi}^{\trsp}(t_i\!\!+\!\!2\delta) \vec{\phi}(t_j\!\!+\!\!2\delta)\!\!-\!\!2\vec{\phi}^{\trsp}(t_i\!\!+\!\!2\delta) \vec{\phi}(t_j\!\!+\!\!\delta)\!\!+\!\!\vec{\phi}^{\trsp}(t_i\!\!+\!\!2\delta)\vec{\phi}(t_j)}{\delta^{4}}\\
&-\frac{2\vec{\phi}^{\trsp}\!(t_i\!\!+\!\!\delta) \vec{\phi}(t_j\!\!+\!\!2\delta)\!\!-\!\!4\vec{\phi}^{\trsp}(t_i\!\!+\!\!\delta) \vec{\phi}(t_j\!\!+\!\!\delta)\!\!+\!\!2\vec{\phi}^{\trsp}\!(t_i\!\!+\!\!\delta) \vec{\phi}(t_j)}{\delta^{4}}\\
&+\frac{\vec{\phi}^{\trsp}(t_i) \vec{\phi}(t_j+2\delta)-2\vec{\phi}^{\trsp}(t_i) \vec{\phi}(t_j+\delta)+\vec{\phi}^{\trsp}(t_i) \vec{\phi}(t_j)}{\delta^{4}}\\
&=\frac{k(t_i+2\delta,t_j+2\delta)-2k(t_i+2\delta,t_j+\delta)+k(t_i+2\delta,t_j)}{\delta^4}\\
&-\frac{2k(t_i+\delta,t_j+2\delta)-4k(t_i+\delta,t_j+\delta)+2k(t_i+\delta,t_j)}{\delta^4}\\
&+\frac{k(t_i,t_j+2\delta)-2k(t_i,t_j+\delta)+k(t_i,t_j)}{\delta^4}.
\end{aligned}
\end{equation*}
Therefore, (\ref{equ:app:basis}) can be kernelized as 
\begin{equation}
\begin{aligned}
\hspace{-0.3cm}\vec{k}(t_i,t_j)
=\left[\begin{matrix}
k_{tt}(t_i,t_j)\vec{I}_{3} \!\!& k_{td}(t_i,t_j)\vec{I}_{3} \!\!\!& k_{ta}(t_i,t_j)\vec{I}_{3} \\
k_{dt}(t_i,t_j)\vec{I}_{3} \!\!& k_{dd}(t_i,t_j)\vec{I}_{3} \!\!\!& k_{da}(t_i,t_j)\vec{I}_{3} \\
k_{at}(t_i,t_j)\vec{I}_{3} \!\!& k_{ad}(t_i,t_j)\vec{I}_{3} \!\!\!& k_{aa}(t_i,t_j)\vec{I}_{3}
\end{matrix}\right]\!\!,
\end{aligned}
\end{equation}
where
\begin{align*}
k_{tt}(t_i,t_j)&\!=\!k(t_i,t_j), \\
k_{td}(t_i,t_j)&\!=\!\big(k(t_i,t_j+\delta)\!\!-\!\!k(t_i,t_j)\big)/\delta, \\
k_{ta}(t_i,t_j)&\!=\!\big(k(t_i,t_j+2\delta)\!\!-\!\!2k(t_i,t_j+\delta)+k(t_i,t_j)\big)/\delta^2,\\
k_{dt}(t_i,t_j)&\!=\!\big(k(t_i+\delta,t_j)-k(t_i,t_j)\big)/\delta, \\
k_{dd}(t_i,t_j)&\!=\!\big(k(t_i+\delta,t_j+\delta)-k(t_i,t_j+\delta)-k(t_i+\delta,t_j)\\&+k(t_i,t_j)\big)/\delta^2, \\
k_{da}(t_i,t_j)&\!=\!\big(k(t_i+\delta,t_j+2\delta)-2k(t_i+\delta,t_j+\delta)\\&+k(t_i+\delta,t_j)-k(t_i,t_j+2\delta)+2k(t_i,t_j+\delta)\\&-k(t_i,t_j)\big)/\delta^3,\\
k_{at}(t_i,t_j)&\!=\!\big(k(t_i+2\delta,t_j)-2k(t_i+\delta,t_j)+k(t_i,t_j)\big)/\delta^2,\\
\end{align*}
\begin{align*}
k_{ad}(t_i,t_j)&\!=\!\big(k(t_i+2\delta,t_j+\delta)-2k(t_i+\delta,t_j+\delta)\\&+k(t_i,t_j+\delta)-k(t_i+2\delta,t_j)+2k(t_i+\delta,t_j)\\&-k(t_i,t_j)\big)/\delta^3,\\
k_{aa}(t_i,t_j)&\!=\!\big(k(t_i+2\delta,t_j+2\delta)-2k(t_i+2\delta,t_j+\delta)\\&+k(t_i+2\delta,t_j)-2k(t_i+\delta,t_j+2\delta)\\&+4k(t_i+\delta,t_j+\delta)-2k(t_i+\delta,t_j)\\&+k(t_i,t_j+2\delta)-2k(t_i,t_j+\delta)+k(t_i,t_j)\big)/\delta^4.
\end{align*}

\end{appendices}


\bibliographystyle{IEEEtran}
\bibliography{bibiography}

\begin{IEEEbiography}[{\includegraphics[width=1in,height=1.25in,trim=15 0 15 0,clip,keepaspectratio]{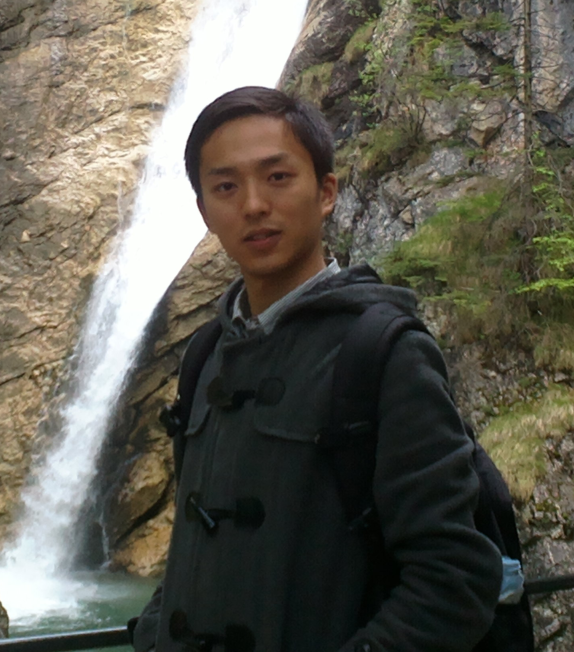}}]{Yanlong Huang} is a university academic fellow at the school of computing, University of Leeds. He received his BSc degree (2008) in Automatic Control and MSc degree (2010) in Control Theory and Control Engineering, both from Nanjing University of Science and Technology, Nanjing, China. After that, he received his PhD degree (2013) in Robotics from the Institute of Automation, Chinese Academy of Sciences, Beijing, China. From 2013 to 2019, he carried out his research as a postdoctoral researcher in Max-Planck Institute for Intelligent Systems and Italian Institute of Technology. His interests include imitation learning, optimal control, reinforcement learning, motion planning and their applications to robotic systems.
\end{IEEEbiography}

\begin{IEEEbiography}[{\includegraphics[width=1in,height=1.25in,clip,keepaspectratio]{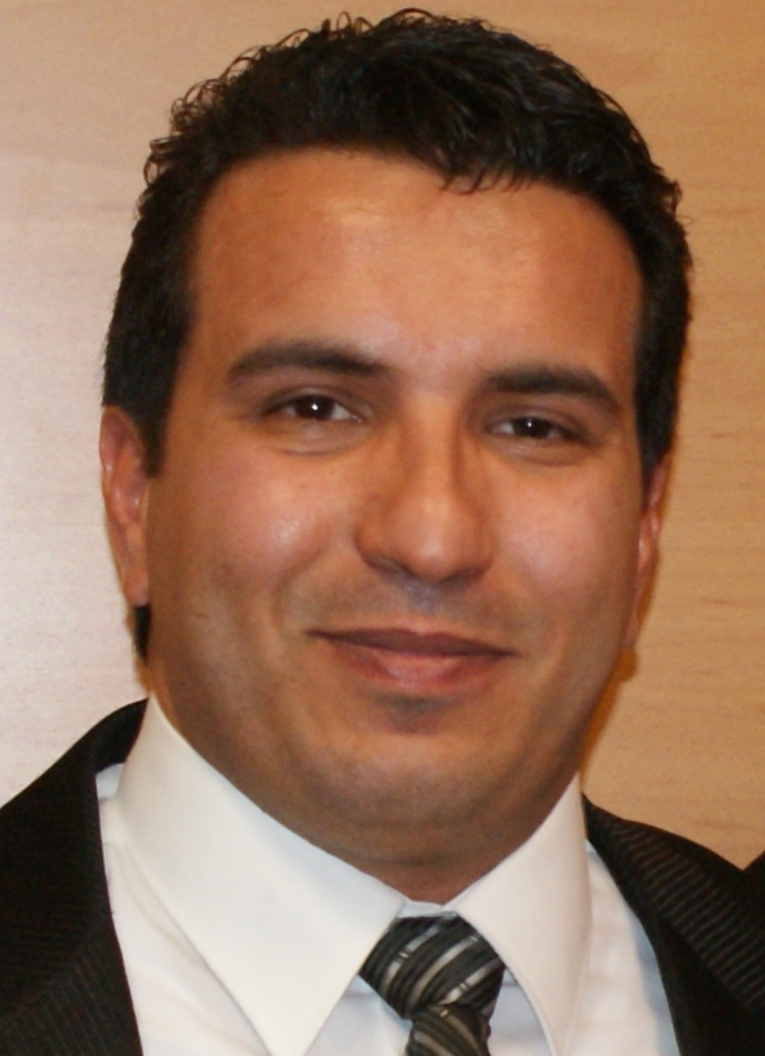}}]{Fares J. Abu-Dakka} received his B.Sc.degree in mechanical engineering from Birzeit University, Palestine, in 2003, and his M.Sc. and Ph.D. degrees in robotics motion planning from the Polytechnic University of Valencia, Spain, in 2006 and 2011, respectively. Currently, he is a senior researcher at Intelligent Robotics Group at the Department of Electrical Engineering and Automation (EEA), Aalto University, Finland. Before that he was researching at ADVR, Istituto Italiano di Tecnologia (IIT). Between 2013 and 2016 he was holding a visiting professor position at the Department of Systems Engineering and Automation of the Carlos III University of Madrid, Spain. His research activities include robot control and learning, human-robot interaction, impedance control, and robot motion planning.
\end{IEEEbiography}

\begin{IEEEbiography}[{\includegraphics[width=1in,height=1.25in,trim=15 0 15 0,clip,keepaspectratio]{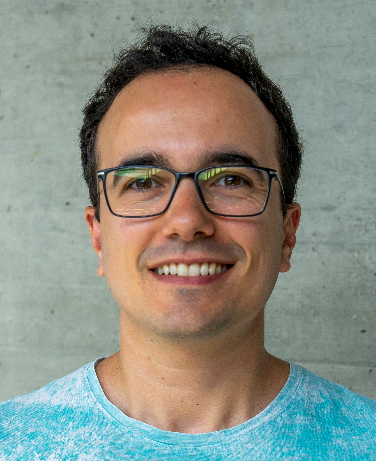}}]{Jo\~ao Silv\'erio} is a postdoctoral researcher at the Idiap Research Institute since July 2019. He received his M.Sc in Electrical and Computer Engineering (2011) from Instituto Superior T\'ecnico (Lisbon, Portugal) and Ph.D in Robotics (2017) from the University of Genoa (Genoa, Italy) and the Italian Institute of Technology, where he was also a postdoctoral researcher until May 2019. He is interested in machine learning for robotics, particularly imitation learning and control. Webpage: \texttt{\footnotesize http://joaosilverio.eu}
\end{IEEEbiography}

\begin{IEEEbiography}[{\includegraphics[width=1in,height=1.25in,clip,keepaspectratio]{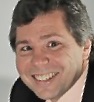}}]{Darwin G. Caldwell} received his BSc. and PhD in Robotics from the University of Hull in 1986 and 1990 respectively. In 1996 he received an MSc in Management from the University of Salford. He is or has been an Honorary Professor at the University of Manchester, the University of Sheffield, the University of Bangor, the Kings College University of London, all in the U.K., and Tianjin University, and Shenzhen Academy of Aerospace Technology in China. The cCub, COMAN, WalkMan, HyQ, HyQ2Max, HyQ-Real and Centauro, Humanoid and quadrupedal robots were all developed in his Department. Prior to this he had worked on the development of the iCub. He is the author or co-author of over 500 academic papers, 20+ patents, and has received over 40 awards and nominations from leading journals and conferences. Prof. Caldwell has been a fellow of the Royal Academy of Engineering since 2015.
\end{IEEEbiography} 

\end{document}